\documentclass[conference]{IEEEtran}
\usepackage{times}

\usepackage[numbers]{natbib}
\usepackage{multicol}
\usepackage[bookmarks=true]{hyperref}
\usepackage[utf8]{inputenc}         %
\usepackage[T1]{fontenc}            %
\usepackage{hyperref}               %
\usepackage{url}                    %
\usepackage{booktabs}               %
\usepackage{amsfonts}               %
\usepackage{nicefrac}               %
\usepackage{microtype}              %
\usepackage[dvipsnames]{xcolor}     %
\usepackage{enumitem}

\usepackage{graphicx}
\usepackage{todonotes}
\usepackage{subcaption}
\usepackage{amsmath}
\usepackage{textcomp, gensymb}
\usepackage{multirow}
\usepackage{algorithm}
\usepackage[noend]{algorithmic}
\usepackage{mathtools}
\usepackage{esvect}
\usepackage{hyperref}
\usepackage{scalerel,xparse}

\DeclarePairedDelimiterX{\infdivx}[2]{(}{)}{%
  #1\;\delimsize\|\;#2%
}

\NewDocumentCommand\emojibeach{}{
    \includegraphics[scale=0.05]{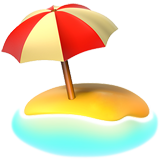}
}

\begin{document}

\title{Learning Fine-Grained Bimanual Manipulation with Low-Cost Hardware}

\author{Tony Z. Zhao$^1$\hspace{3mm}
Vikash Kumar$^3$\hspace{3mm}
Sergey Levine$^2$\hspace{3mm}
Chelsea Finn$^1$\\
$^1$ Stanford University $^2$ UC Berkeley $^3$ Meta 
}

\makeatletter
\let\@oldmaketitle\@maketitle%
\renewcommand{\@maketitle}{\@oldmaketitle%
\includegraphics[width=18.1cm,trim={0.3cm 44.35cm 0.3cm 0.5cm},clip]{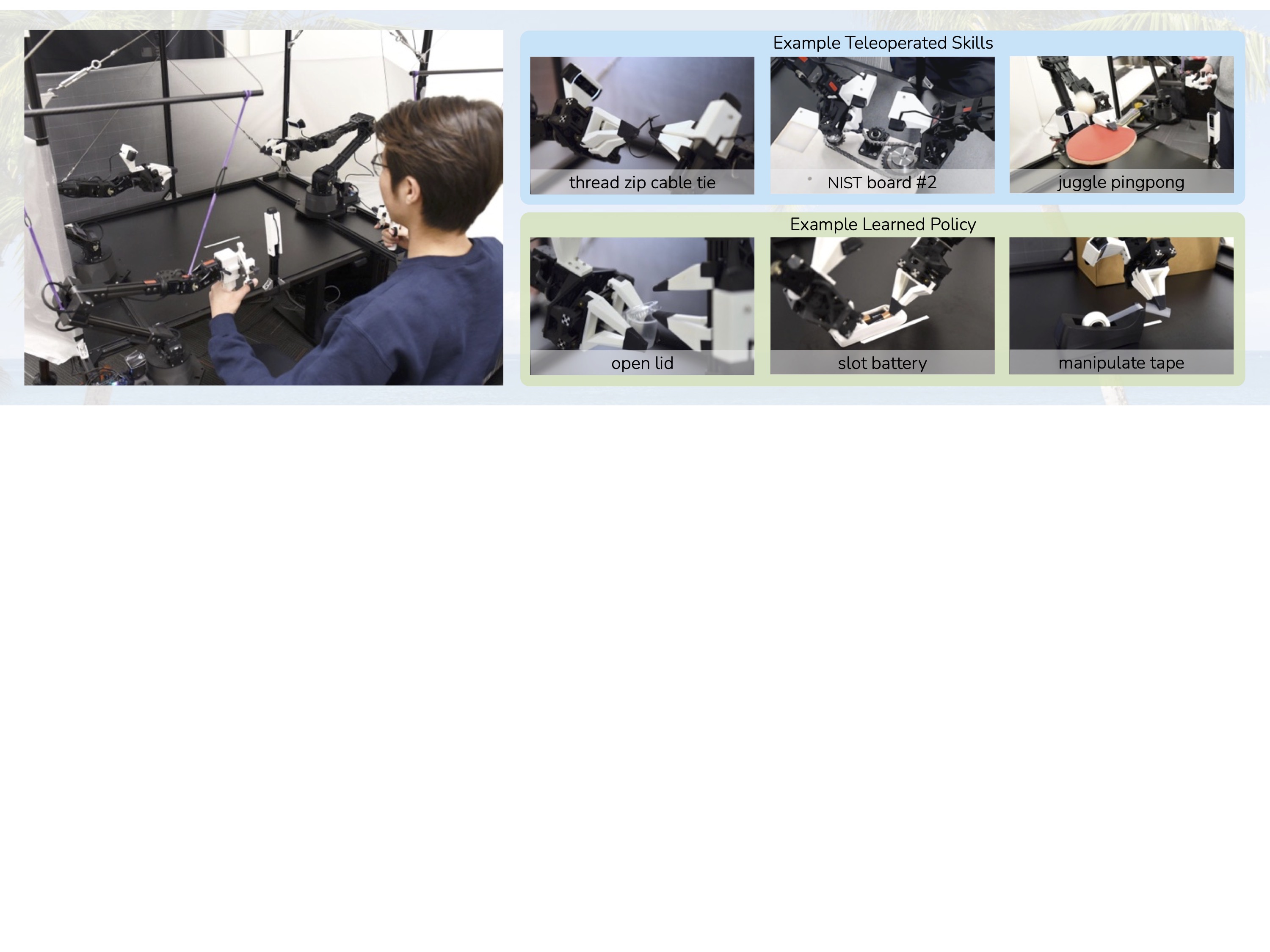}
\vspace{-0.65cm}
\captionof{figure}{\small\textit{ALOHA\emojibeach: \underline{A} \underline{L}ow-cost \underline{O}pen-source \underline{Ha}rdware System for Bimanual Teleoperation}. The whole system costs <\$20k with off-the-shelf robots and 3D printed components. \textit{Left:} The user teleoperates by backdriving the leader robots, with the follower robots mirroring the motion. \textit{Right:} \textit{ALOHA} is capable of precise, contact-rich, and dynamic tasks. We show examples of both teleoperated and learned skills.}
\label{fig:aloha}
\vspace{-0.35cm}
}
    
\makeatother

\maketitle

\begin{abstract}
Fine manipulation tasks, such as threading cable ties or slotting a battery, are notoriously difficult for robots because they require precision, careful coordination of contact forces, and closed-loop visual feedback.
Performing these tasks typically requires high-end robots, accurate sensors, or careful calibration, which can be expensive and difficult to set up.
\textit{Can learning enable low-cost and imprecise hardware to perform these fine manipulation tasks?}
We present a low-cost system that performs end-to-end imitation learning directly from real demonstrations, collected with a custom teleoperation interface. %
Imitation learning, however, presents its own challenges, particularly in high-precision domains: %
errors in the policy can compound over time, %
and human demonstrations can be non-stationary.
To address these challenges, we develop a simple yet novel algorithm, \textit{Action Chunking with Transformers (ACT)}, %
which learns a generative model over action sequences.
ACT allows the robot to learn 6 difficult tasks in the real world, such as opening a translucent condiment cup and slotting a battery with 80-90\% success, with only 10 minutes worth of demonstrations. %
Project website: \href{https://tonyzhaozh.github.io/aloha/}{tonyzhaozh.github.io/aloha}
\end{abstract}

\IEEEpeerreviewmaketitle

\section{Introduction}
Fine manipulation tasks involve precise, closed-loop feedback and require high degrees of hand-eye coordination to adjust and re-plan in response to changes in the environment. 
Examples of such manipulation tasks include opening the lid of a condiment cup or slotting a battery, which involve delicate operations such as pinching, prying, and tearing rather than broad-stroke motions such as picking and placing.
Take opening the lid of a condiment cup in Figure~\ref{fig:aloha} as an example, where the cup is initialized upright on the table: the right gripper needs to first tip it over, and nudge it into the opened left gripper. Then the left gripper closes gently and lifts the cup off the table. Next, one of the right fingers approaches the cup from below and pries the lid open.
Each of these steps requires high precision, delicate hand-eye coordination, and rich contact. Millimeters of error would lead to task failure.

Existing systems for fine manipulation use expensive robots and high-end sensors for precise state estimation \cite{ke2021grasping, Sundaresan2021UntanglingDN, Kim2022RobotPB, Paradis2020IntermittentVS}.
In this work, we seek to develop a low-cost system for fine manipulation that is, in contrast, accessible and reproducible.
However, low-cost hardware is inevitably less precise than high-end platforms, making the sensing and planning challenge more pronounced.
One promising direction to resolve this is to incorporate learning into the system.
Humans also do not have industrial-grade proprioception \cite{HORVATH2022}, and yet we are able to perform delicate tasks by learning from closed-loop visual feedback and actively compensating for errors.
In our system, we therefore train an end-to-end policy that directly maps RGB images from commodity web cameras to the actions.
This pixel-to-action formulation is particularly suitable for fine manipulation, because fine manipulation often involves objects with complex physical properties, such that learning the manipulation policy is much simpler than modeling the whole environment.
Take the condiment cup example: modeling the contact when nudging the cup, and also the deformation when prying open the lid involves complex physics on a large number of degrees of freedom.
Designing a model accurate enough for planning would require significant research and task specific engineering efforts.
In contrast, the policy of nudging and opening the cup is much simpler, since a closed-loop policy can react to different positions of the cup and lid rather than precisely anticipating how it will move in advance.

Training an end-to-end policy, however, presents its own challenges.
The performance of the policy depends heavily on the training data distribution, and in the case of fine manipulation, high-quality human demonstrations can provide tremendous value by allowing the system to learn from human dexterity.
We thus build a low-cost yet dexterous teleoperation system for data collection, and a novel imitation learning algorithm that learns effectively from the demonstrations. %
We overview each component in the following two paragraphs.

\noindent \textbf{Teleoperation system.}
We devise a teleoperation setup %
with two sets of low-cost, off-the-shelf robot arms. They are approximately scaled versions of each other, and we use joint-space mapping for teleoperation.
We augment this setup with 3D printed components for easier backdriving, leading to a highly capable teleoperation system within a \$20k budget.
We showcase its capabilities in Figure~\ref{fig:aloha}, including teleoperation of precise tasks such as threading a zip tie, dynamic tasks such as juggling a ping pong ball, and contact-rich tasks such as assembling the chain in the NIST board \#2 \cite{nist_2022}.

\noindent \textbf{Imitation learning algorithm.}
Tasks that require precision and visual feedback present a significant challenge for imitation learning, even with high-quality demonstrations.
Small errors in the predicted action can incur large differences in the state, exacerbating the ``compounding error'' problem of imitation learning \cite{Ross2010ARO, Tu2021OnTS, ke2021grasping}. %
To tackle this, we take inspiration from \textit{action chunking}, a concept in psychology that describes how sequences of actions are grouped together as a chunk, and executed as one unit~\cite{lai_huang_gershman_2022}.
In our case, the policy predicts the target joint positions for the next $k$ timesteps, rather than just one step at a time.
This reduces the effective horizon of the task by $k$-fold, mitigating compounding errors.
Predicting action sequences also helps tackle temporally correlated confounders \cite{Swamy2022CausalIL}, such as pauses in demonstrations that are hard to model with Markovian single-step policies.
To further improve the smoothness of the policy, we propose \textit{temporal ensembling}, which queries the policy more frequently and averages across the overlapping action chunks. 
We implement action chunking policy with Transformers \cite{Vaswani2017AttentionIA}, an architecture designed for sequence modeling, and train it as a conditional VAE (CVAE) \cite{Sohn2015LearningSO, Kingma2013AutoEncodingVB} to capture the variability in human data.
We name our method \textit{Action Chunking with Transformers} (ACT), and find that it significantly outperforms previous imitation learning algorithms on a range of simulated and real-world fine manipulation tasks.

The key contribution of this paper is a low-cost system for learning fine manipulation, comprising a teleoperation system and a novel imitation learning algorithm.
The teleoperation system, despite its low cost, enables tasks with high precision and rich contacts.
The imitation learning algorithm, Action Chunking with Transformers (ACT), is capable of learning precise, close-loop behavior and drastically outperforms previous methods.
The synergy between these two parts allows learning of 6 fine manipulation skills directly in the real-world, such as opening a translucent condiment cup and slotting a battery with 80-90\% success, from only 10 minutes or 50 demonstration trajectories.

\section{Related Work}

\noindent \textbf{Imitation learning for robotic manipulation.}
Imitation learning allows a robot to directly learn from experts.
Behavioral cloning (BC) \cite{Pomerleau1988ALVINNAA} is one of the simplest imitation learning algorithms, casting imitation as supervised learning from observations to actions.
Many works have then sought to improve BC, for example by incorporating history with various architectures \cite{Mandlekar2021WhatMI, Shafiullah2022BehaviorTC, Jang2022BCZZT, Brohan2022RT1RT},
using a different training objective \cite{Florence2021ImplicitBC, pari2021surprising}, and including regularization \cite{Rahmatizadeh2017VisionBasedMM}.
Other works emphasize the multi-task or few-shot aspect of imitation learning~\cite{Duan2017OneShotIL, James2018TaskEmbeddedCN, Dasari2020TransformersFO},
leveraging language \cite{Shridhar2021CLIPortWA, Shridhar2022PerceiverActorAM, Jang2022BCZZT, Brohan2022RT1RT}, or exploiting the specific task structure \cite{Pastor2009LearningAG, Zeng2020TransporterNR, Johns2021CoarsetoFineIL, Shridhar2022PerceiverActorAM}.
Scaling these imitation learning algorithms with more data has led to impressive systems that can generalize to new objects, instructions, or scenes \cite{Ebert2021BridgeDB, Jang2022BCZZT, Brohan2022RT1RT, Kim2022RobotPB}.
In this work, we focus on building an imitation learning system that is low-cost yet capable of performing delicate, fine manipulation tasks. We tackle this from both hardware and software, by building a high-performance teleoperation system, and a novel imitation learning algorithm that drastically improves previous methods on fine manipulation tasks.

\begin{figure*}[t!]
\centering
\includegraphics[width=\textwidth,trim={4cm 0.8cm 18cm 4.5cm},clip]{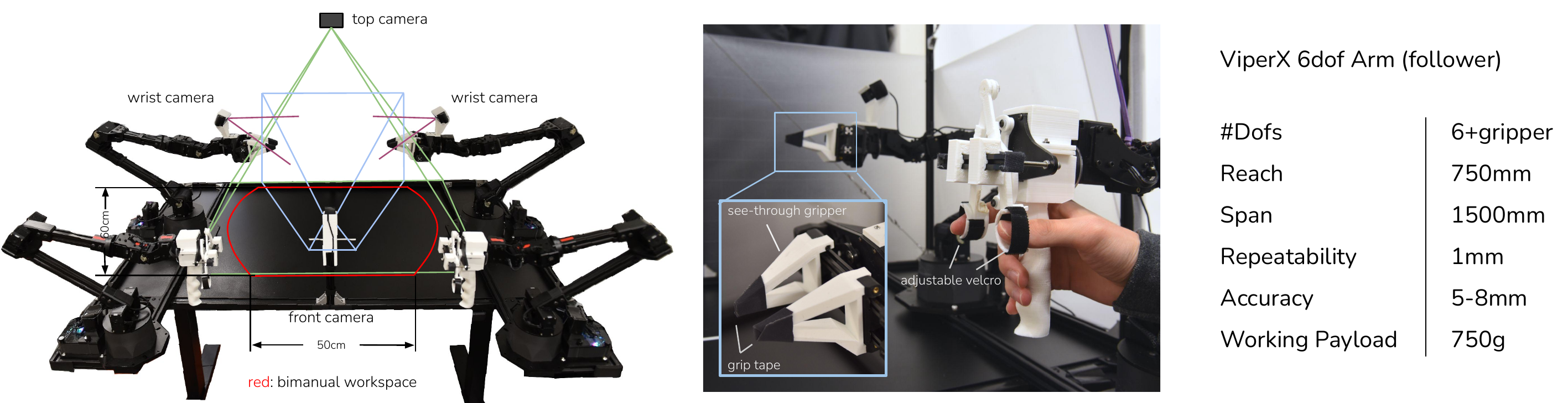}
\caption{\small \textit{Left:} Camera viewpoints of the front, top, and two wrist cameras, together with an illustration of the bimanual workspace of \textit{ALOHA}. \textit{Middle:} Detailed view of the ``handle and scissor'' mechanism and custom grippers. \textit{Right:} Technical spec of the ViperX 6dof robot \cite{vx}.}
\vspace{-0.5cm}
\label{fig:setup_close}
\end{figure*}

\noindent \textbf{Addressing compounding errors.}
\label{compounding_errors}
A major shortcoming of BC is compounding errors, where errors from previous timesteps accumulate and cause the robot to drift off of its training distribution, leading to hard-to-recover states~\cite{Ross2010ARO, Tu2021OnTS}. 
This problem is particularly prominent in the fine manipulation setting \cite{ke2021grasping}.
One way to mitigate compounding errors is to allow additional on-policy interactions and expert corrections, such as DAgger \cite{Ross2010ARO} and its variants \cite{Kelly2018HGDAggerII, Menda2018EnsembleDAggerAB, Hoque2021ThriftyDAggerBN}.
However, expert annotation can be time-consuming and unnatural with a teleoperation interface \cite{ke2021grasping}.
One could also inject noise at demonstration collection time to obtain datasets with corrective behavior \cite{Laskey2017DARTNI}, but for fine manipulation, such noise injection can directly lead to task failure, reducing the dexterity of teleoperation system.
To circumvent these issues, previous works generate synthetic correction data in an offline manner \cite{Florence2019SelfSupervisedCI, ke2021grasping, Zhou2023NeRFIT}.
While they are limited to settings where low-dimensional states are available, or 
a specific type of task like grasping.
Due to these limitations, we need to address the compounding error problem from a different angle, compatible with high-dimensional visual observations.
We propose to reduce the effective horizon of tasks through action chunking, i.e., predicting an action sequence instead of a single action, and then ensemble across overlapping action chunks to produce trajectories that are both accurate and smooth.

\noindent \textbf{Bimanual manipulation.}
Bimanual manipulation has a long history in robotics, and has gained popularity with the lowering of hardware costs.
Early works tackle bimanual manipulation from a classical control perspective, with known environment dynamics \cite{Smith2012DualAM, Salehian2018AUF}, but designing such models can be time-consuming, and they may not be accurate for objects with complex physical properties.
More recently, learning has been incorporated into bimanual systems, such as reinforcement learning \cite{Chen2022TowardsHB, Chitnis2019EfficientBM}, imitating human demonstrations \cite{Kroemer2015TowardsLH, Lee2015LearningFM, Stepputtis2022ASF, Xie2020DeepIL, Kim2022RobotPB}, or learning to predict key points that chain together motor primitives \cite{Ha2021FlingBotTU, Grannen2020UntanglingDK, Shivakumar2022SGTM2A}.
Some of the works also focus on fine-grained manipulation tasks such as knot untying, cloth flattening, or even threading a needle \cite{Grannen2020UntanglingDK, Ganapathi2020LearningDV, Kim2021GazeBasedDR}, while using robots that are considerably more expensive, e.g. the da Vinci surgical robot or ABB YuMi.
Our work turns to low-cost hardware, e.g. arms that cost around \$5k each, and seeks to enable them to perform high-precision, closed-loop tasks.
Our teleoperation setup is most similar to \citet{Kim2022RobotPB}, which also uses joint-space mapping between the leader and follower robots.
Unlike this previous system, we do not make use of special encoders, sensors, or machined components. We build our system with only off-the-shelf robots and a handful of 3D printed parts, allowing non-experts to assemble it in less than 2 hours.

\section{ALOHA: \underline{A} \underline{L}ow-cost \underline{O}pen-source \underline{Ha}rdware System for Bimanual Teleoperation}

We seek to develop an accessible and high-performance teleoperation system for fine manipulation. We summarize our design considerations into the following 5 principles.
\begin{enumerate}
    \item \textbf{Low-cost}: The entire system should be within budget for most robotic labs, comparable to a single industrial arm.
    \item \textbf{Versatile}: It can be applied to a wide range of fine manipulation tasks with real-world objects. 
    \item \textbf{User-friendly}: The system should be intuitive, reliable, and easy to use.
    \item \textbf{Repairable}: The setup can be easily repaired by researchers, when it inevitably breaks.
    \item \textbf{Easy-to-build}: It can be quickly assembled by researchers, with easy-to-source materials.
\end{enumerate}

\begin{figure*}[t!]
    \centering
    \vspace*{-7mm}
    \includegraphics[width=17cm,trim={1cm 11.5cm 0 0},clip]{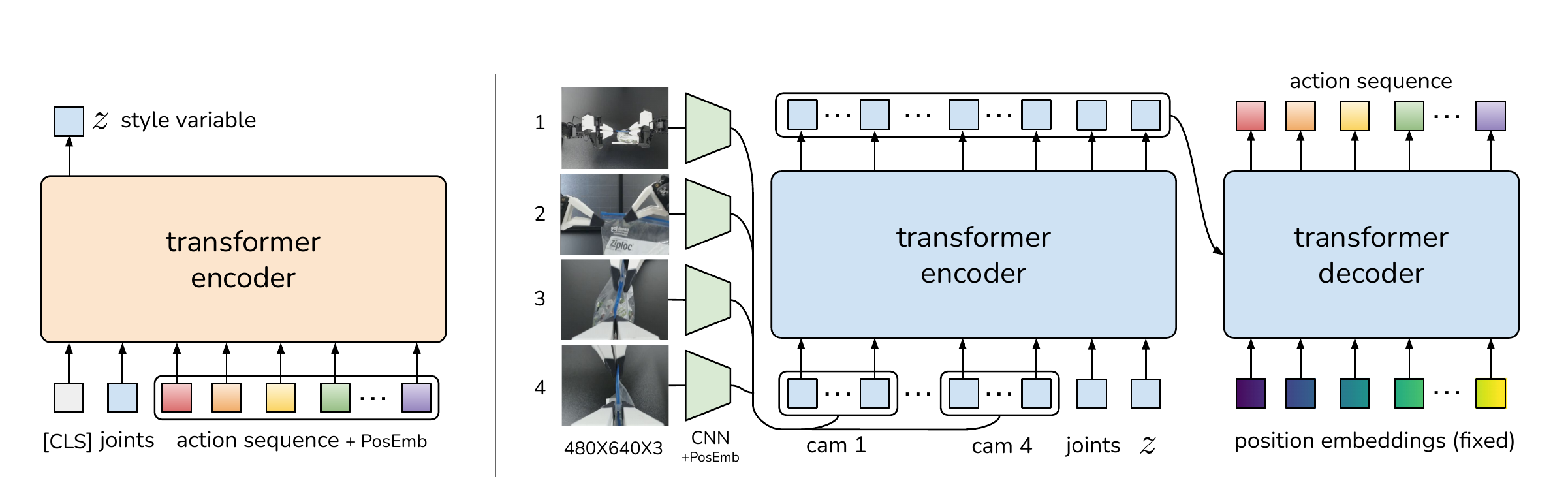}
    \vspace*{-3mm}
    \caption{\small \textit{Architecture of Action Chunking with Transformers (ACT)}. We train ACT as a Conditional VAE (CVAE), which has an encoder and a decoder. \textit{Left:} The encoder of the CVAE compresses action sequence and joint observation into $z$, the style variable. The encoder is discarded at test time.
    \textit{Right:} The decoder or policy of ACT synthesizes images from multiple viewpoints, joint positions, and $z$ with a transformer encoder, and predicts a sequence of actions with a transformer decoder. $z$ is simply set to the mean of the prior (i.e. zero) at test time. 
    }
    \label{fig:act}
    \vspace*{-5mm}
\end{figure*}

When choosing the robot to use, principles 1, 4, and 5 lead us to build a bimanual parallel-jaw grippers setup with two ViperX 6-DoF robot arms \cite{vx, Wiznitzer_interbotix_ros_manipulators}. We do not employ dexterous hands due to price and maintenance considerations.
The ViperX arm used has a working payload of 750g and 1.5m span, with an accuracy of 5-8mm.
The robot is modular and simple to repair: in the case of motor failure, the low-cost Dynamixel motors can be easily replaced.
The robot can be purchased off-the-shelf for around \$5600.
The OEM fingers, however, are not versatile enough to handle fine manipulation tasks. We thus design our own 3D printed ``see-through'' fingers 
and fit it with gripping tape (Fig~\ref{fig:setup_close}). 
This allows for good visibility when performing delicate operations, and robust grip even with thin plastic films.

We then seek to design a teleoperation system that is maximally user-friendly around the ViperX robot. Instead of mapping the hand pose captured by a VR controller or camera to the end-effector pose of the robot, i.e. task-space mapping, we use direct joint-space mapping from a smaller robot, WidowX, manufactured by the same company and costs \$3300 \cite{wx}.
The user teleoperates by backdriving the smaller WidowX (``the leader''), whose joints are synchronized with the larger ViperX (``the follower'').
When developing the setup, we noticed a few benefits of using joint-space mapping compared to task-space.
(1) Fine manipulation often requires operating near singularities of the robot, which in our case has 6 degrees of freedom and no redundancy. Off-the-shelf inverse kinematics (IK) fails frequently in this setting. Joint space mapping, on the other hand, guarantees high-bandwidth control within the joint limits, while also requiring less computation and reducing latency.
(2) The weight of the leader robot prevents the user from moving too fast, and also dampens small vibrations. %
We notice better performance on precise tasks with joint-space mapping rather than holding a VR controller.
To further improve the teleoperation experience, we design a 3D-printed ``handle and scissor'' mechanism that can be retrofitted to the leader robot (Fig~\ref{fig:setup_close}).
It reduces the force required from the operator to backdrive the motor, and allows for continuous control of the gripper, instead of binary opening or closing.
We also design a rubber band load balancing mechanism 
that partially counteracts the gravity on the leader side. It reduces the effort needed from the operator and makes longer teleoperation sessions (e.g. >30 minutes) possible. 
We include more details about the setup in the \href{https://tonyzhaozh.github.io/aloha/}{project website}.

The rest of the setup includes a robot cage with 20$\times$20mm aluminum extrusions, reinforced by crossing steel cables. There is a total of four Logitech C922x webcams, each streaming 480$\times$640 RGB images. Two of the webcams are mounted on the wrist of the follower robots, allowing for a close-up view of the grippers. The remaining two cameras are mounted on the front and at the top respectively (Fig~\ref{fig:setup_close}).
Both the teleoperation and data recording happen at 50Hz.

With the design considerations above, we build the bimanual teleoperation setup \textit{ALOHA} within a 20k USD budget, comparable to a single research arm such as Franka Emika Panda.
\textit{ALOHA} enables the teleoperation of:
\begin{itemize}[leftmargin=*]

    \item \textbf{Precise tasks} such as threading zip cable ties, picking credit cards out of wallets, and opening or closing ziploc bags.

    \item \textbf{Contact-rich tasks} such as inserting 288-pin RAM into a computer motherboard, turning pages of a book, and assembling the chains and belts in the NIST board \#2 \cite{nist_2022}

    \item \textbf{Dynamic tasks} such as juggling a ping pong ball with a real ping pong paddle, balancing the ball without it falling off, and swinging open plastic bags in the air.
\end{itemize}

\noindent Skills such as threading a zip tie, inserting RAM, and juggling ping pong ball, to our knowledge, are not available for existing teleoperation systems with 5-10x the budget \cite{Handa2019DexPilotVT, shadow_robot_2022}. We include a more detailed price \& capability comparison in Appendix~\ref{appendix_compare}, as well as more skills that \textit{ALOHA} is capable of in Figure~\ref{fig:more_teleop}.
To make \textit{ALOHA} more accessible, we open-source all software and hardware with a detailed tutorial covering 3D printing, assembling the frame to software installations. %
You can find the tutorial on the \href{https://tonyzhaozh.github.io/aloha/}{project website}.

\begin{figure}[t]
    \centering
    \includegraphics[width=0.65\linewidth,trim={1.5cm 2.5cm 16cm 10cm},clip]{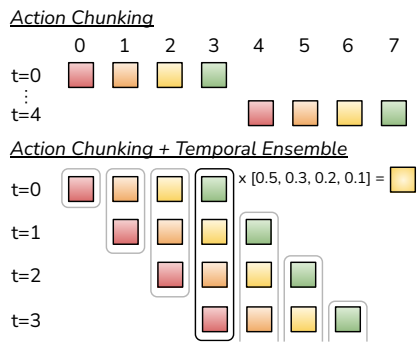}
    \vspace*{-3mm}
    \caption{\small We employ both Action Chunking and Temporal Ensembling when applying actions, instead of interleaving observing and executing.
    }
    \label{fig:ta}
    \vspace*{-5mm}
\end{figure}

\section{Action Chunking with Transformers}
As we will see in Section~\ref{experiments}, existing imitation learning algorithms perform poorly on fine-grained tasks that require high-frequency control and closed-loop feedback. We therefore develop a novel algorithm, \textit{Action Chunking with Transformers (ACT)}, to leverage the data collected by \textit{ALOHA}.
We first summarize the pipeline of training ACT, then dive into each of the design choices.

To train ACT on a new task, we first collect human demonstrations using \textit{ALOHA}. We record the joint positions of the leader robots (i.e. input from the human operator) and use them as actions. 
It is important to use the leader joint positions instead of the follower's, because the amount of force applied is implicitly defined by the difference between them, through the low-level PID controller.
The observations are composed of the current joint positions of follower robots and the image feed from 4 cameras.
Next, we train ACT to predict the \textit{sequence of future actions} given the current observations. 
An action here corresponds to the target joint positions for both arms in the next time step.
Intuitively, ACT tries to imitate what a human operator would do in the following time steps given current observations.
These target joint positions are then tracked by the low-level, high-frequency PID controller inside Dynamixel motors.
At test time, we load the policy that achieves the lowest validation loss and roll it out in the environment.
The main challenge that arises is compounding errors, where errors from previous actions lead to states that are outside of training distribution.

\subsection{Action Chunking and Temporal Ensemble} %

To combat the compounding errors of imitation learning in a way that is compatible with pixel-to-action
policies (Figure~\ref{compounding_errors}), we seek to reduce the effective horizon of long trajectories collected at high frequency.
We are inspired by \textit{action chunking}, a neuroscience concept where individual actions are grouped together and executed as one unit, making them more efficient to store and execute \cite{lai_huang_gershman_2022}.
Intuitively, a chunk of actions could correspond to grasping a corner of the candy wrapper or inserting a battery into the slot.
In our implementation, we fix the chunk size to be $k$: every $k$ steps, the agent receives an observation, generates the next $k$ actions, and executes the actions in sequence (Figure~\ref{fig:ta}).
This implies a $k$-fold reduction in the effective horizon of the task.
Concretely, the policy models $\pi_\theta(a_{t:t+k} | s_t)$ instead of $\pi_\theta(a_t| s_t)$.
Chunking can also help model non-Markovian behavior in human demonstrations.
Specifically, a single-step policy would struggle with temporally correlated confounders, such as pauses in the middle of a demonstration \cite{Swamy2022CausalIL}, since the behavior not only depends on the state, but also the timestep.
Action chunking can mitigate this issue when the confounder is within a chunk, without introducing the causal confusion issue for history-conditioned policies \cite{Haan2019CausalCI}.

A na\"{i}ve implementation of action chunking can be sub-optimal: a new environment observation is incorporated abruptly every $k$ steps and can result in jerky robot motion.
To improve smoothness and avoid discrete switching between executing and observing, we query the policy at every timestep.
This makes different action chunks overlap with each other, and at a given timestep there will be more than one predicted action. We illustrate this in Figure~\ref{fig:ta} and propose a \textit{temporal ensemble} to combine these predictions.
Our \textit{temporal ensemble} performs a weighted average over these predictions with an exponential weighting scheme $w_{i} = \exp(-m*i)$, where $w_0$ is the weight for the oldest action.
The speed for incorporating new observation is governed by $m$, where a smaller $m$ means faster incorporation.
We note that unlike typical smoothing, where the current action is aggregated with actions in adjacent timesteps, which leads to bias, we aggregate actions predicted for the \textit{same} timestep. 
This procedure also incurs no additional training cost, only extra inference-time computation.
In practice, we find both action chunking and temporal ensembling to be important for the success of ACT, which produces precise and smooth motion. We discuss these components in more detail in the ablation studies in Subsection~\ref{ablate_ac_ta}.

\begin{algorithm}[t]
\small
  	\caption{ACT Training}
  	\label{alg:act_train}
  	\begin{algorithmic}[1]
  	\STATE Given: Demo dataset $\mathcal{D}$, chunk size $k$, weight $\beta$.
        \STATE Let $a_t$, $o_t$ represent action and observation at timestep $t$, $\Bar{o}_t$ represent $o_t$ without image observations.
  	\STATE Initialize encoder $q_\phi(z | a_{t:t+k}, \Bar{o}_t)$ 
        \STATE Initialize decoder $\pi_\theta(\hat{a}_{t:t+k} | o_t, z)$
  	\FOR{iteration $n=1, 2, ...$}
  	    \STATE Sample $o_t$, $a_{t:t+k}$ from $\mathcal{D}$
            \STATE Sample $z$ from $q_\phi(z | a_{t:t+k}, \Bar{o}_t)$
            \STATE Predict $\hat{a}_{t:t+k}$ from $\pi_\theta(\hat{a}_{t:t+k} | o_t, z)$
            \STATE $\mathcal{L}_{reconst} = \mathit{MSE}(\hat{a}_{t:t+k}, a_{t:t+k})$
            \STATE $\mathcal{L}_{reg} = D_{\mathit{KL}}\infdivx{q_\phi(z | a_{t:t+k}, \Bar{o}_t)}{\mathcal{N}(0,\mathit{I})}$
            \STATE Update $\theta$, $\phi$ with ADAM and $\mathcal{L} = \mathcal{L}_{reconst} + \beta\mathcal{L}_{reg}$
  	\ENDFOR
  	\end{algorithmic}
\end{algorithm}
\begin{algorithm}[t]
\small
  	\caption{ACT Inference}
  	\label{alg:act_inference}
  	\begin{algorithmic}[1]
  	\STATE Given: trained $\pi_\theta$, episode length $T$, weight $m$.
        \STATE Initialize FIFO buffers $\mathcal{B}[0:T]$, where $\mathcal{B}[t]$ stores actions predicted \textit{for} timestep $t$.
  	\FOR{timestep $t=1, 2, ... T$}
  	    \STATE Predict $\hat{a}_{t:t+k}$ with $\pi_\theta(\hat{a}_{t:t+k} | o_t, z)$ where $z=0$
            \STATE Add $\hat{a}_{t:t+k}$ to buffers $\mathcal{B}[t:t+k]$ respectively
            \STATE Obtain current step actions $A_t = \mathcal{B}[t]$
            \STATE Apply $a_t = \sum_{i} w_i A_t[i]/\sum_{i} w_i$, with $w_{i} = \exp(-m*i)$
  	\ENDFOR
  	\end{algorithmic}
\end{algorithm}

\subsection{Modeling human data}
Another challenge that arises is learning from noisy human demonstrations.
Given the same observation, a human can use different trajectories to solve the task. 
Humans will also be more stochastic in regions where precision matters less \cite{Li2006OptimalCF}.
Thus, it is important for the policy to focus on regions where high precision matters.
We tackle this problem by training our action chunking policy as a generative model.
Specifically, we train the policy as a conditional variational autoencoder (CVAE) \cite{Sohn2015LearningSO}, to generate an action sequence conditioned on current observations.
The CVAE has two components: a CVAE encoder and a CVAE decoder, illustrated on the left and right side of Figure~\ref{fig:act} respectively.
The CVAE encoder only serves to train the CVAE decoder (the policy) and is discarded at test time.
Specifically, the CVAE encoder predicts the mean and variance of the style variable $z$'s distribution, which is parameterized as a diagonal Gaussian, given the current observation and action sequence as inputs. 
For faster training in practice, we leave out the image observations and only condition on the proprioceptive observation and the action sequence.
The CVAE decoder, i.e. the policy, conditions on both $z$ and the current observations (images + joint positions) to predict the action sequence.
At test time, we set $z$ to be the mean of the prior distribution i.e. zero to deterministically decode.
The whole model is trained to maximize the log-likelihood of demonstration action chunks, i.e. $\min_\theta - \sum_{s_t, a_{t:t+k} \in D} \log \pi_\theta( a_{t:t+k} | s_t)$, with the standard VAE objective which has two terms: a reconstruction loss and a term that regularizes the encoder to a Gaussian prior.
Following~\cite{Higgins2016betaVAELB}, we weight the second term with a hyperparameter $\beta$.
Intuitively, higher $\beta$ will result in less information transmitted in $z$ \cite{Tishby2015DeepLA}.
Overall, we found the CVAE objective to be essential in learning precise tasks from human demonstrations. We include a more detailed discussion in Subsection~\ref{ablate_vae}.

\subsection{Implementing ACT}
We implement the CVAE encoder and decoder with transformers, as transformers are designed for both synthesizing information across a sequence and generating new sequences.
The CVAE encoder is implemented with a BERT-like transformer encoder \cite{Devlin2019BERTPO}.
The inputs to the encoder are the current joint positions and the target action sequence of length $k$ from the demonstration dataset, prepended by a learned ``[CLS]'' token similar to BERT. This forms a $k+2$ length input (Figure \ref{fig:act} left).
After passing through the transformer, the feature corresponding to ``[CLS]'' is used to predict the mean and variance of the ``style variable'' $z$, which is then used as input to the decoder.
The CVAE decoder (i.e. the policy) takes the current observations and $z$ as the input, and predicts the next $k$ actions (Figure \ref{fig:act} right). We use ResNet image encoders, a transformer encoder, and a transformer decoder to implement the CVAE decoder. Intuitively, the transformer encoder synthesizes information from different camera viewpoints, the joint positions, and the style variable, and the transformer decoder generates a coherent action sequence.
The observation includes 4 RGB images, each at $480\times640$ resolution, and joint positions for two robot arms (7+7=14 DoF in total).
The action space is the absolute joint positions for two robots, a 14-dimensional vector.
Thus with action chunking, the policy outputs a $k\times 14$ tensor given the current observation.
The policy first process the images with ResNet18 backbones \cite{He2015DeepRL}, which convert $480\times640\times3$ RGB images into $15\times20\times512$ feature maps. We then flatten along the spatial dimension to obtain a sequence of $300\times512$. To preserve the spatial information, we add a 2D sinusoidal position embedding to the feature sequence \cite{Carion2020EndtoEndOD}.
Repeating this for all 4 images gives a feature sequence of $1200\times512$ in dimension.
We then append two more features: the current joint positions and the ``style variable'' $z$. They are projected from their original dimensions to $512$ through linear layers respectively.
Thus, the input to the transformer encoder is $1202\times512$.
The transformer decoder conditions on the encoder output through cross-attention, 
where the input sequence is a fixed position embedding, with dimensions $k \times 512$, and the keys and values are coming from the encoder.
This gives the transformer decoder an output dimension of $k \times 512$, which is then down-projected with an MLP into $k \times 14$, corresponding to the predicted target joint positions for the next $k$ steps. 
We use L1 loss for reconstruction instead of the more common L2 loss: we noted that L1 loss leads to more precise modeling of the action sequence.
We also noted degraded performance when using delta joint positions as actions instead of target joint positions.
We include a detailed architecture diagram in Appendix~\ref{appendix_architecture}.

We summarize the training and inference of ACT in Algorithms \ref{alg:act_train} and \ref{alg:act_inference}. The model has around 80M parameters, and we train from scratch for each task. The training takes around 5 hours on a single 11G RTX 2080 Ti GPU, and the inference time is around 0.01 seconds on the same machine.

\begin{figure*}[htp]
    \centering
    \includegraphics[width=17cm,trim={0 8cm 0 1cm},clip]{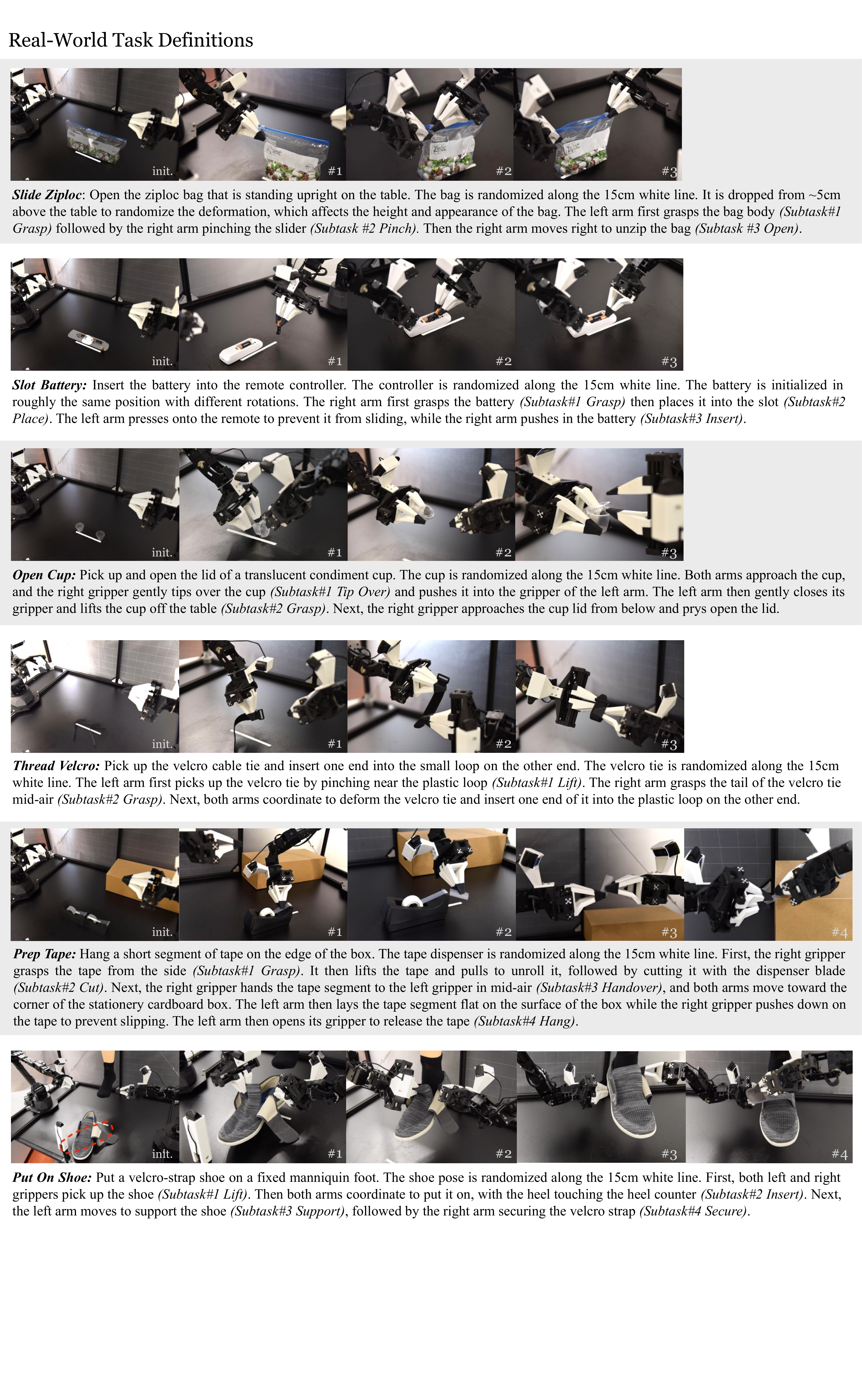}
    \caption{\small \textit{Real-World Task Definitions.} For each of the 6 real-world tasks, we illustrate the initializations and the subtasks.}
    \label{fig:real_tasks}
\end{figure*}

\begin{figure*}[htp]
    \centering
    \includegraphics[width=18cm,trim={0 18.3cm 0 1cm},clip]{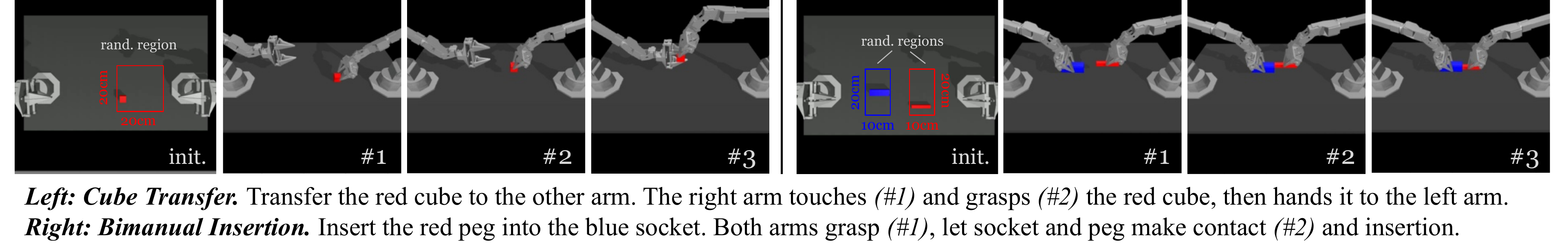}
    \caption{\small \textit{Simulated Task Definitions.} For each of the 2 simulated tasks, we illustrate the initializations and the subtasks.}
    \label{fig:sim_tasks}
\end{figure*}

\begin{table*}[t!]
\centering
\setlength{\tabcolsep}{8pt}
\begin{tabular}{lllllllcccccc} %
\toprule
 &  \multicolumn{3}{c}{\textbf{Cube Transfer} (sim)} & \multicolumn{3}{c}{\textbf{Bimanual Insertion} (sim)} & \multicolumn{3}{c}{\textbf{Slide Ziploc} (real)} & \multicolumn{3}{c}{\textbf{Slot Battery} (real)} \\ 
\cmidrule(lr){2-4}
\cmidrule(lr){5-7}
\cmidrule(lr){8-10}
\cmidrule(lr){11-13}
& Touched & Lifted & Transfer & Grasp & Contact & Insert & Grasp & Pinch & Open & Grasp & Place & Insert \\
\midrule
BC-ConvMLP & 34 | 3 & 17 | 1 & {\hspace{4pt}1} | 0 & {\hspace{4pt}5} | 0 & {\hspace{4pt}1} | 0 & {\hspace{4pt}1} | 0 & 0 & 0 & 0 & 0 & 0 & 0 \\
\addlinespace
BeT        & 60 | 16 & 51 | 13 & 27 | 1 & 21 | 0 & {\hspace{4pt}4} | 0 & {\hspace{4pt}3} | 0 & 8 & 0 & 0 & 4 & 0 & 0 \\
\addlinespace
RT-1       & 44 | 4 & 33 | 2 & {\hspace{4pt}2} | 0 & {\hspace{4pt}2} | 0 & {\hspace{4pt}0} | 0 & {\hspace{4pt}1} | 0 & 4 & 0 & 0 & 4 & 0 & 0 \\
\addlinespace
VINN       & 13 | 17 & {\hspace{4pt}9} | 11 & {\hspace{4pt}3} | 0 & {\hspace{4pt}6} | 0 & {\hspace{4pt}1} | 0 & {\hspace{4pt}1} | 0 & 28 & 0 & 0 & 20 & 0 & 0 \\
\addlinespace
ACT (Ours)  & \textbf{97 | 82} & \textbf{90 | 60} & \textbf{86 | 50} & \textbf{93 | 76} & \textbf{90 | 66} & \textbf{32 | 20} & \textbf{92} & \textbf{96} & \textbf{88} & \textbf{100} & \textbf{100} & \textbf{96} \\

\bottomrule
\end{tabular}
\caption{\small Success rate (\%) for 2 simulated and 2 real-world tasks, comparing our method with 4 baselines. For the two simulated tasks, we report [training with scripted data | training with human data], with 3 seeds and 50 policy evaluations each. For the real-world tasks, we report training with human data, with 1 seed and 25 evaluations. Overall, ACT significantly outperforms previous methods.}
\label{table:main_results}
\end{table*}

\begin{table*}[t!]
\centering
\setlength{\tabcolsep}{6.2pt}
\begin{tabular}{ccccccccccccccc}
\toprule
 &  \multicolumn{3}{c}{\textbf{Open Cup} (real)} & \multicolumn{3}{c}{\textbf{Thread Velcro} (real)} & \multicolumn{4}{c}{\textbf{Prep Tape} (real)} & \multicolumn{4}{c}{\textbf{Put On Shoe} (real)} \\ 
\cmidrule(lr){2-4}
\cmidrule(lr){5-7}
\cmidrule(lr){8-11}
\cmidrule(lr){12-15}
& Tip Over & Grasp & Open Lid & Lift & Grasp & Insert & Grasp & Cut & Handover & Hang & Lift & Insert & Support & Secure \\
\midrule
\addlinespace
BeT & 12 & 0 & 0 & 24 & 0 & 0 & 8 & 0 & 0 & 0 & 12 & 0 & 0 & 0 \\
\addlinespace
ACT (Ours)  & \textbf{100} & \textbf{96} & \textbf{84} & \textbf{92} & \textbf{40} & \textbf{20} & \textbf{96} & \textbf{92} & \textbf{72} & \textbf{64} & \textbf{100} & \textbf{92} & \textbf{92} & \textbf{92}  \\

\bottomrule
\end{tabular}
\caption{\small Success rate (\%) for the remaining 3 real-world tasks. We only compare with the best performing baseline BeT.}
\vspace{-0.5cm}
\label{table:main_results2}
\end{table*}

\section{Experiments}
\label{experiments}

We present experiments to evaluate ACT's performance on fine manipulation tasks.
For ease of reproducibility, we build two simulated fine manipulation tasks in MuJoCo \cite{Todorov2012MuJoCoAP}, in addition to 6 real-world tasks with \textit{ALOHA}. We provide videos for each task on the \href{https://tonyzhaozh.github.io/aloha/}{project website}.

\subsection{Tasks}
All 8 tasks require fine-grained, bimanual manipulation, and are illustrated in Figure~\ref{fig:real_tasks}.
For \textbf{\textit{Slide Ziploc}}, the right gripper needs to accurately grasp the slider of the ziploc bag and open it, with the left gripper securing the body of the bag.
For \textbf{\textit{Slot Battery}}, the right gripper needs to first place the battery into the slot of the remote controller, then using the tip of fingers to delicately push in the edge of the battery, until it is fully inserted. Because the spring inside the battery slot causes the remote controller to move in the opposite direction during insertion, the left gripper pushes down on the remote to keep it in place.
For \textbf{\textit{Open Cup}}, the goal is to open the lid of a small condiment cup. Because of the cup's small size, the grippers cannot grasp the body of the cup by just approaching it from the side. Therefore we leverage both grippers: the right fingers first lightly tap near the edge of the cup to tip it over, and then nudge it into the open left gripper. This nudging step requires high precision and closing the loop on visual perception. The left gripper then closes gently and lifts the cup off the table, followed by the right finger prying open the lid, which also requires precision to not miss the lid or damage the cup.
The goal of \textbf{\textit{Thread Velcro}} is to insert one end of a velcro cable tie into the small loop attached to other end. The left gripper needs to first pick up the velcro tie from the table, followed by the right gripper pinching the tail of the tie in mid-air. Then, both arms coordinate to insert one end of the velcro tie into the other in mid-air. The loop measures 3mm x 25mm, while the velcro tie measures 2mm x 10-25mm depending on the position. For this task to be successful, the robot must use visual feedback to correct for perturbations with each grasp,
as even a few millimeters of error during the first grasp will compound in the second grasp mid-air, giving more than a 10mm deviation in the insertion phase.
For \textbf{\textit{Prep Tape}}, the goal is to hang a small segment of the tape on the edge of a cardboard box. The right gripper first grasps the tape and cuts it with the tape dispenser's blade, and then hands the tape segment to the left gripper mid-air. Next, both arms approach the box, the left arm gently lays the tape segment on the box surface, and the right fingers push down on the tape to prevent slipping, followed by the left arm opening its gripper to release the tape. Similar to \textit{Thread Velcro}, this task requires multiple steps of delicate coordination between the two arms.
For \textbf{\textit{Put On Shoe}}, the goal is to put the shoe on a fixed manniquin foot, and secure it with the shoe's velcro strap. The arms would first need to grasp the tongue and collar of the shoe respectively, lift it up and approach the foot. Putting the shoe on is challenging because of the tight fitting: the arms would need to coordinate carefully to nudge the foot in, and both grasps need to be robust enough to counteract the friction between the sock and shoe. Then, the left arm goes around to the bottom of the shoe to support it from dropping, followed by the right arm flipping the velcro strap and pressing it against the shoe to secure. The task is only considered successful if the shoe clings to the foot after both arms releases.
For the simulated task \textbf{\textit{Transfer Cube}}, the right arm needs to first pick up the red cube lying on the table, then place it inside the gripper of the other arm. Due to the small clearance between the cube and the left gripper (around 1cm), small errors could result in collisions and task failure.
For the simulated task \textbf{\textit{Bimanual Insertion}}, the left and right arms need to pick up the socket and peg respectively, and then insert in mid-air so the peg touches the ``pins'' inside the socket. The clearance is around 5mm in the insertion phase.
For all 8 tasks, the initial placement of the objects is either varied randomly along the 15cm white reference line (real-world tasks), or uniformly in 2D regions (simulated tasks). We provide illustrations of both the initial positions and the subtasks in Figure~\ref{fig:real_tasks} and \ref{fig:sim_tasks}. Our evaluation will additionally report the performance for each of these subtasks.

In addition to the delicate bimanual control required to solve these tasks, the objects we use also present a significant perception challenge.
For example, the ziploc bag is largely transparent, with a thin blue sealing line. Both the wrinkles on the bag and the reflective candy wrappers inside can vary during the randomization, and distract the perception system.
Other transparent or translucent objects include the tape and both the lid and body of the condiment cup, making them hard to perceive precisely and ill-suited for depth cameras.
The black table top also creates a low-contrast against many objects of interest, such as the black velcro cable tie and the black tape dispenser. Especially from the top view, it is challenging to localize the velcro tie because of the small projected area.

\subsection{Data Collection}
For all 6 real-world tasks, we collect demonstrations using \textit{ALOHA} teleoperation. 
Each episode takes 8-14 seconds for the human operator to perform depending on the complexity of the task, which translates to 400-700 time steps given the control frequency of 50Hz.
We record 50 demonstrations for each task, except for \textit{Thread Velcro} which has 100.
The total amount for demonstrations is thus around 10-20 minutes of data for each task, and 30-60 minutes in wall-clock time because of resets and teleoperator mistakes.
For the two simulated tasks, we collect two types of demonstrations: one type with a scripted policy and one with human demonstrations. To teleoperate in simulation, we use the ``leader robots'' of \textit{ALOHA} to control the simulated robot, with the operator looking at the real-time renderings of the environment on the monitor.
In both cases, we record 50 successful demonstrations.

We emphasize that all human demonstrations are inherently stochastic, even though a single person collects all of the demonstrations.
Take the mid-air hand handover of the tape segment as an example: the exact position of the handover is different across each episode. The human has no visual or haptic reference to perform it in the same position. Thus to successfully perform the task, the policy will need to learn that the two grippers should never collide with each other during the handover, and the left gripper should always move to a position that can grasp the tape, instead of trying to memorize where exactly the handover happens, which can vary across demonstrations.

\subsection{Experiment Results}
\label{expr_results}
We compare ACT with four prior imitation learning methods.
\textbf{\textit{BC-ConvMLP}} is the simplest yet most widely used baseline~\cite{Zhang2017DeepIL,Jang2022BCZZT}, which processes the current image observations with a convolutional network, whose output features are concatenated with the joint positions to predict the action.
\textbf{\textit{BeT}} \cite{Shafiullah2022BehaviorTC} also leverages Transformers as the architecture, but with key differences: (1) no action chunking: the model predicts one action given the history of observations; and (2) the image observations are pre-processed by a separately trained frozen visual encoder. That is, the perception and control networks are not jointly optimized.
\textbf{\textit{RT-1}} \cite{Brohan2022RT1RT} is another Transformer-based architecture that predicts one action from a fixed-length history of past observations.
Both \textit{BeT} and \textit{RT-1} discretize the action space: the output is a categorical distribution over discrete bins, but with an added continuous offset from the bin-center in the case of \textit{BeT}. Our method, ACT, instead directly predicts continuous actions, motivated by the precision required in fine manipulation.
Lastly, \textbf{\textit{VINN}} \cite{pari2021surprising} is a non-parametric method that assumes access to the demonstrations at test time. Given a new observation, it retrieves the $k$ observations with the most similar visual features, and returns an action using weighted $k$-nearest-neighbors. The visual feature extractor is a pretrained ResNet finetuned on demonstration data with unsupervised learning.
We carefully tune the hyperparameters of these four prior methods using cube transfer. Details of the hyperparameters are provided in Appendix~\ref{appendix_hparams}.

As a detailed comparison with prior methods, we report the average success rate in Table~\ref{table:main_results} for two simulated and two real tasks.
For simulated tasks, we average performance across 3 random seeds with 50 trials each. We report the success rate on both scripted data (left of separation bar) and human data (right of separation bar).
For real-world tasks, we run one seed and evaluate with 25 trials. %
ACT achieves the highest success rate compared to all prior methods, outperforming the second best algorithm by a large margin on each task.
For the two simulated tasks with scripted or human data, ACT outperforms the best previous method in success rate by 59\%, 49\%, 29\%, and 20\%.
While previous methods are able to make progress in the first two subtasks, the final success rate remains low, below 30\%.
For the two real-world tasks \textit{Slide Ziploc} and \textit{Slot Battery}, ACT achieves 88\% and 96\% final success rates respectively, with other methods making no progress past the first stage.
We attribute the poor performance of prior methods to compounding errors and non-Markovian behavior in the data: the behavior degrades significantly towards the end of an episode, and the robot can pause indefinitely for certain states.
ACT mitigates both issues with action chunking. Our ablations in Subsection~\ref{ablate_ac_ta} also shows that chunking can significantly improve these prior methods when incorporated. 
In addition, we notice a drop in performance for all methods when switching from scripted data to human data in simulated tasks: the stochasticity and multi-modality of human demonstrations make imitation learning a lot harder.

\begin{figure*}[t!]
    \centering
    \includegraphics[width=18.7cm,trim={1cm 6.5cm 0 11cm},clip]{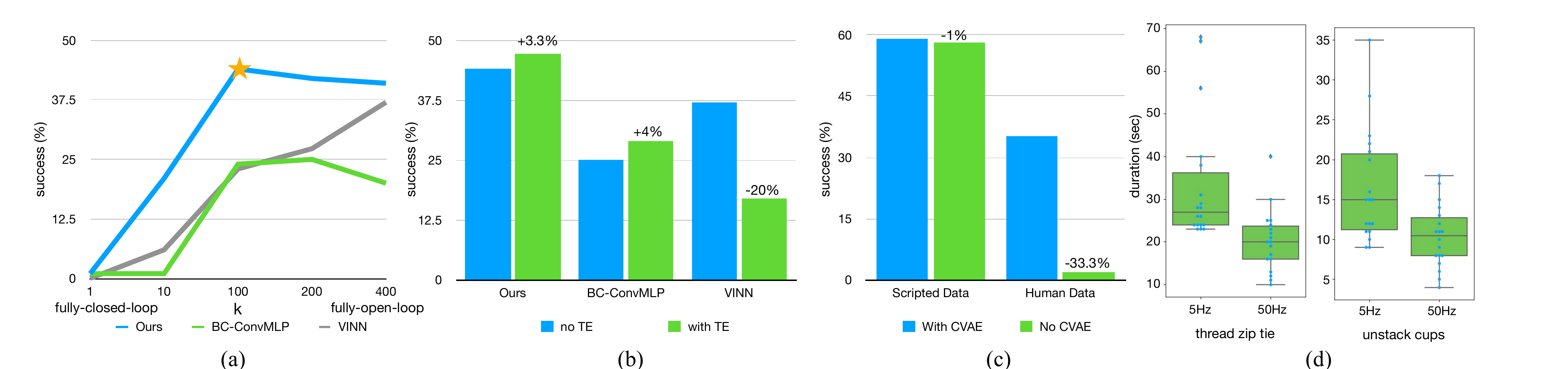}
    \vspace*{-5mm}
    \caption{\small \textit{(a)} We augment two baselines with action chunking, with different values of chunk size $k$ on the x-axis, and success rate on the y-axis. Both methods significantly benefit from action chunking, suggesting that it is a generally useful technique. \textit{(b)} Temporal Ensemble (TE) improves our method and \textit{BC-ConvMLP}, while hurting \textit{VINN}. \textit{(c)} We compare with and without the CVAE training, showing that it is crucial when learning from human data. \textit{(d)} We plot the distribution of task completion time in our user study, where we task participants to perform two tasks, at 5Hz or 50Hz teleoperation frequency. Lowering the frequency results in a 62\% slowdown in completion time.}
    \label{fig:ablation}
    \vspace*{-5mm}
\end{figure*}

We report the success rate of the 3 remaining real-world tasks in Table~\ref{table:main_results2}. For these tasks, we only compare with BeT, which has the highest task success rate so far.
Our method ACT reaches 84\% success for \textit{Cup Open}, 20\% for \textit{Thread Velcro}, 64\% for \textit{Prep Tape} and 92\% for \textit{Put On Shoe}, again outperforming BeT, which achieve zero final success on these challenging tasks.
We observe relatively low success of ACT in \textit{Thread Velcro}, where the success rate decreased by roughly half at every stage, from 92\% success at the first stage to 20\% final success.
The failure modes we observe are 1) at stage 2, the right arm closes its gripper too early and fails to grasp the tail of the cable tie mid-air, and 2) in stage 3, the insertion is not precise enough and misses the loop. 
In both cases, it is hard to determine the exact position of the cable tie from image observations: the contrast is low between the black cable tie and the background, and the cable tie only occupies a small fraction of the image. We include examples of image observations in Appendix~\ref{appendix_example_image}.

\section{Ablations}
ACT employs action chunking and temporal ensembling to mitigate compounding errors and better handle non-Markovian demonstrations. It also trains the policy as a conditional VAE to model the noisy human demonstrations.
In this section, we ablate each of these components, together with a user study that highlights the necessity of high-frequency control in \textit{ALOHA}.
We report results across a total of four settings: two simulated tasks with scripted or human demonstration.

\subsection{Action Chunking and Temporal Ensembling}
\label{ablate_ac_ta}

In Subsection~\ref{expr_results},
we observed that ACT significantly outperforms previous methods that only predict single-step actions, with the hypothesis that action chunking is the key design choice.
Since $k$ dictates how long the sequence in each ``chunk'' is, we can analyze this hypothesis by varying $k$. $k=1$ corresponds to no action chunking, and $k=episode\_length$ corresponds to fully open-loop control, where the robot outputs the entire episode's action sequence based on the first observation. We disable temporal ensembling in these experiments to only measure the effect of chunking, and trained separate policies for each $k$.
In Figure~\ref{fig:ablation} (a), we plot the success rate averaged across 4 settings, corresponding to 2 simulated tasks with either human or scripted data, with the blue line representing ACT without the temporal ensemble.
We observe that performance improves drastically from 1\% at $k=1$ to 44\% at $k=100$, then slightly tapers down with higher $k$.
This illustrates that more chunking and a lower effective horizon generally improve performance. We attribute the slight dip at $k=200, 400$ (i.e., close to open-loop control) to the lack of reactive behavior and the difficulty in modeling long action sequences.
To further evaluate the effectiveness and generality of action chunking, we augment two baseline methods with action chunking. For \textit{BC-ConvMLP}, we simply increase the output dimension to $k * action\_dim$, and for \textit{VINN}, we retrieve the next $k$ actions.
We visualize their performance in Figure~\ref{fig:ablation} (a) with different $k$, showing trends consistent with ACT, where more action chunking improves performance.
While ACT still outperforms both augmented baselines with sizable gains, these results suggest that action chunking is generally beneficial for imitation learning in these settings. 

We then ablate the temporal ensemble by comparing the highest success rate with or without it, again across the 4 aforementioned tasks and different $k$.
We note that experiments with and without the temporal ensemble are separately tuned: hyperparameters that work best for no temporal ensemble may not be optimal with a temporal ensemble.
In Figure~\ref{fig:ablation} (b), we show that
\textit{BC-ConvMLP} benefits from temporal ensembling the most with a 4\% gain, followed by a 3.3\% gain for our method.
We notice a performance drop for \textit{VINN}, a non-parametric method. We hypothesize that a temporal ensemble mostly benefits parametric methods by smoothing out the modeling errors. In contrast, \textit{VINN} retrieves ground-truth actions from the dataset and does not suffer from this issue.

\subsection{Training with CVAE}
\label{ablate_vae}

We train ACT with CVAE objective to model human demonstrations, which can be noisy and contain multi-modal behavior.
In this section, we compare with ACT without the CVAE objective, which simply predicts a sequence of actions given current observation, and trained with L1 loss.
In Figure~\ref{fig:ablation} (c), we visualize the success rate aggregated across 2 simulated tasks, and separately plot training with scripted data and with human data.
We can see that when training on scripted data, the removal of CVAE objective makes almost no difference in performance, because dataset is fully deterministic.
While for human data, there is a significant drop from 35.3\% to 2\%.
This illustrates that the CVAE objective is crucial when learning from human demonstrations.

\subsection{Is High-Frequency Necessary?}

Lastly, we conduct a user study to illustrate the necessity of high-frequency teleoperation for fine manipulation.
With the same hardware setup, we lower the frequency from 50Hz to 5Hz, a control frequency that is similar to recent works that use high-capacity deep networks for imitation learning \cite{Brohan2022RT1RT, Zhou2023NeRFIT}.
We pick two fine-grained tasks: threading a zip cable tie and un-stacking two plastic cups. Both require millimeter-level precision and closed-loop visual feedback.
We perform the study with 6 participants who have varying levels of experience with teleoperation, though none had used \textit{ALOHA} before.
The participants were recruited from among computer science graduate students, with 4 men and 2 women aged 22-25
The order of tasks and frequencies are randomized for each participant, and each participant was provided with a 2 minutes practice period before each trial.
We recorded the time it took to perform the task for 3 trials, and visualize the data in Figure~\ref{fig:ablation} (d).
On average, it took 33s for participants to thread the zip tie at 5Hz, which is lowered to 20s at 50Hz. For separating plastic cups, increasing the control frequency lowered the task duration from 16s to 10s.
Overall, our setup (i.e. 50Hz) allows the participants to perform highly dexterous and precise tasks in a short amount of time.
However, reducing the frequency from 50Hz to 5Hz results in a 62\% increase in teleoperation time. 
We then use ``Repeated Measures Designs'', a statistical procedure, to formally verify that 50Hz teleoperation outperforms 5Hz with p-value <0.001. We include more details about the study in Appendix~\ref{appendix_user_study}.

\section{Limitations and Conclusion}
We present a low-cost system for fine manipulation, comprising a teleoperation system \textit{ALOHA} and a novel imitation learning algorithm \textit{ACT}.
The synergy between these two parts allows us to learn fine manipulation skills directly in the real-world, such as opening a translucent condiment cup and slotting a battery with a 80-90\% success rate and around 10 min of demonstrations. 
While the system is quite capable, there exist tasks that are beyond the capability of either the robots or the learning algorithm, such as buttoning up a dress shirt. We include a more detailed discussion about limitations in Appendix~\ref{appendix_limitations}. Overall, we hope that this low-cost open-source system represents an important step and accessible resource towards advancing fine-grained robotic manipulation.

\section*{Acknowledgement}
We thank members of the IRIS lab at Stanford for their support and feedback. We also thank Siddharth Karamcheti, Toki Migimatsu, Staven Cao, Huihan Liu, Mandi Zhao, Pete Florence and Corey Lynch for helpful discussions. Tony Zhao is supported by Stanford Robotics Fellowship sponsored by FANUC, in addition to Schmidt Futures and ONR Grant N00014-21-1-2685.

\bibliographystyle{plainnat}
\bibliography{references}

\begin{thebibliography}{71}
\providecommand{\natexlab}[1]{#1}
\providecommand{\url}[1]{\texttt{#1}}
\expandafter\ifx\csname urlstyle\endcsname\relax
  \providecommand{\doi}[1]{doi: #1}\else
  \providecommand{\doi}{doi: \begingroup \urlstyle{rm}\Url}\fi

\bibitem[vx()]{vx}
Viperx 300 robot arm 6dof.
\newblock URL
  \url{https://www.trossenrobotics.com/viperx-300-robot-arm-6dof.aspx}.

\bibitem[wx()]{wx}
Widowx 250 robot arm 6dof.
\newblock URL
  \url{https://www.trossenrobotics.com/widowx-250-robot-arm-6dof.aspx}.

\bibitem[you(2014)]{youtube_2014}
Highly dexterous manipulation system - capabilities - part 1, Nov 2014.
\newblock URL \url{https://www.youtube.com/watch?v=TearcKVj0iY}.

\bibitem[nis(2022)]{nist_2022}
Assembly performance metrics and test methods, Apr 2022.
\newblock URL
  \url{https://www.nist.gov/el/intelligent-systems-division-73500/robotic-grasping-and-manipulation-assembly/assembly}.

\bibitem[sha(2022)]{shadow_robot_2022}
Teleoperated robots - shadow teleoperation system, Nov 2022.
\newblock URL \url{https://www.shadowrobot.com/teleoperation/}.

\bibitem[Arunachalam et~al.(2022)Arunachalam, G{\"u}zey, Chintala, and
  Pinto]{arunachalam2022holo}
Sridhar~Pandian Arunachalam, Irmak G{\"u}zey, Soumith Chintala, and Lerrel
  Pinto.
\newblock Holo-dex: Teaching dexterity with immersive mixed reality.
\newblock \emph{arXiv preprint arXiv:2210.06463}, 2022.

\bibitem[Brohan et~al.(2022)Brohan, Brown, Carbajal, Chebotar, Dabis, Finn,
  Gopalakrishnan, Hausman, Herzog, Hsu, Ibarz, Ichter, Irpan, Jackson,
  Jesmonth, Joshi, Julian, Kalashnikov, Kuang, Leal, Lee, Levine, Lu, Malla,
  Manjunath, Mordatch, Nachum, Parada, Peralta, Perez, Pertsch, Quiambao, Rao,
  Ryoo, Salazar, Sanketi, Sayed, Singh, Sontakke, Stone, Tan, Tran, Vanhoucke,
  Vega, Vuong, Xia, Xiao, Xu, Xu, Yu, and Zitkovich]{Brohan2022RT1RT}
Anthony Brohan, Noah Brown, Justice Carbajal, Yevgen Chebotar, Joseph Dabis,
  Chelsea Finn, Keerthana Gopalakrishnan, Karol Hausman, Alexander Herzog,
  Jasmine Hsu, Julian Ibarz, Brian Ichter, Alex Irpan, Tomas Jackson, Sally
  Jesmonth, Nikhil~J. Joshi, Ryan~C. Julian, Dmitry Kalashnikov, Yuheng Kuang,
  Isabel Leal, Kuang-Huei Lee, Sergey Levine, Yao Lu, Utsav Malla, Deeksha
  Manjunath, Igor Mordatch, Ofir Nachum, Carolina Parada, Jodilyn Peralta,
  Emily Perez, Karl Pertsch, Jornell Quiambao, Kanishka Rao, Michael~S. Ryoo,
  Grecia Salazar, Pannag~R. Sanketi, Kevin Sayed, Jaspiar Singh, Sumedh~Anand
  Sontakke, Austin Stone, Clayton Tan, Huong Tran, Vincent Vanhoucke, Steve
  Vega, Quan~Ho Vuong, F.~Xia, Ted Xiao, Peng Xu, Sichun Xu, Tianhe Yu, and
  Brianna Zitkovich.
\newblock Rt-1: Robotics transformer for real-world control at scale.
\newblock \emph{ArXiv}, abs/2212.06817, 2022.

\bibitem[Carion et~al.(2020)Carion, Massa, Synnaeve, Usunier, Kirillov, and
  Zagoruyko]{Carion2020EndtoEndOD}
Nicolas Carion, Francisco Massa, Gabriel Synnaeve, Nicolas Usunier, Alexander
  Kirillov, and Sergey Zagoruyko.
\newblock End-to-end object detection with transformers.
\newblock \emph{ArXiv}, abs/2005.12872, 2020.

\bibitem[Chen et~al.(2022)Chen, Yang, Wu, Wang, Feng, Jiang, McAleer, Dong, Lu,
  and Zhu]{Chen2022TowardsHB}
Yuanpei Chen, Yaodong Yang, Tianhao Wu, Shengjie Wang, Xidong Feng, Jiechuan
  Jiang, Stephen McAleer, Hao Dong, Zongqing Lu, and Song-Chun Zhu.
\newblock Towards human-level bimanual dexterous manipulation with
  reinforcement learning.
\newblock \emph{ArXiv}, abs/2206.08686, 2022.

\bibitem[Chitnis et~al.(2019)Chitnis, Tulsiani, Gupta, and
  Gupta]{Chitnis2019EfficientBM}
Rohan Chitnis, Shubham Tulsiani, Saurabh Gupta, and Abhinav~Kumar Gupta.
\newblock Efficient bimanual manipulation using learned task schemas.
\newblock \emph{2020 IEEE International Conference on Robotics and Automation
  (ICRA)}, pages 1149--1155, 2019.

\bibitem[Dasari and Gupta(2020)]{Dasari2020TransformersFO}
Sudeep Dasari and Abhinav~Kumar Gupta.
\newblock Transformers for one-shot visual imitation.
\newblock In \emph{Conference on Robot Learning}, 2020.

\bibitem[de~Haan et~al.(2019)de~Haan, Jayaraman, and Levine]{Haan2019CausalCI}
Pim de~Haan, Dinesh Jayaraman, and Sergey Levine.
\newblock Causal confusion in imitation learning.
\newblock In \emph{Neural Information Processing Systems}, 2019.

\bibitem[Devlin et~al.(2019)Devlin, Chang, Lee, and
  Toutanova]{Devlin2019BERTPO}
Jacob Devlin, Ming-Wei Chang, Kenton Lee, and Kristina Toutanova.
\newblock Bert: Pre-training of deep bidirectional transformers for language
  understanding.
\newblock \emph{ArXiv}, abs/1810.04805, 2019.

\bibitem[Duan et~al.(2017)Duan, Andrychowicz, Stadie, Ho, Schneider, Sutskever,
  Abbeel, and Zaremba]{Duan2017OneShotIL}
Yan Duan, Marcin Andrychowicz, Bradly~C. Stadie, Jonathan Ho, Jonas Schneider,
  Ilya Sutskever, P.~Abbeel, and Wojciech Zaremba.
\newblock One-shot imitation learning.
\newblock \emph{ArXiv}, abs/1703.07326, 2017.

\bibitem[Ebert et~al.(2021)Ebert, Yang, Schmeckpeper, Bucher, Georgakis,
  Daniilidis, Finn, and Levine]{Ebert2021BridgeDB}
Frederik Ebert, Yanlai Yang, Karl Schmeckpeper, Bernadette Bucher, Georgios
  Georgakis, Kostas Daniilidis, Chelsea Finn, and Sergey Levine.
\newblock Bridge data: Boosting generalization of robotic skills with
  cross-domain datasets.
\newblock \emph{ArXiv}, abs/2109.13396, 2021.

\bibitem[Florence et~al.(2019)Florence, Manuelli, and
  Tedrake]{Florence2019SelfSupervisedCI}
Peter~R. Florence, Lucas Manuelli, and Russ Tedrake.
\newblock Self-supervised correspondence in visuomotor policy learning.
\newblock \emph{IEEE Robotics and Automation Letters}, 5:\penalty0 492--499,
  2019.

\bibitem[Florence et~al.(2021)Florence, Lynch, Zeng, Ramirez, Wahid, Downs,
  Wong, Lee, Mordatch, and Tompson]{Florence2021ImplicitBC}
Peter~R. Florence, Corey Lynch, Andy Zeng, Oscar Ramirez, Ayzaan Wahid, Laura
  Downs, Adrian~S. Wong, Johnny Lee, Igor Mordatch, and Jonathan Tompson.
\newblock Implicit behavioral cloning.
\newblock \emph{ArXiv}, abs/2109.00137, 2021.

\bibitem[Ganapathi et~al.(2020)Ganapathi, Sundaresan, Thananjeyan, Balakrishna,
  Seita, Grannen, Hwang, Hoque, Gonzalez, Jamali, Yamane, Iba, and
  Goldberg]{Ganapathi2020LearningDV}
Aditya Ganapathi, Priya Sundaresan, Brijen Thananjeyan, Ashwin Balakrishna,
  Daniel Seita, Jennifer Grannen, Minho Hwang, Ryan Hoque, Joseph Gonzalez,
  Nawid Jamali, Katsu Yamane, Soshi Iba, and Ken Goldberg.
\newblock Learning dense visual correspondences in simulation to smooth and
  fold real fabrics.
\newblock \emph{2021 IEEE International Conference on Robotics and Automation
  (ICRA)}, pages 11515--11522, 2020.

\bibitem[Grannen et~al.(2020)Grannen, Sundaresan, Thananjeyan, Ichnowski,
  Balakrishna, Hwang, Viswanath, Laskey, Gonzalez, and
  Goldberg]{Grannen2020UntanglingDK}
Jennifer Grannen, Priya Sundaresan, Brijen Thananjeyan, Jeffrey Ichnowski,
  Ashwin Balakrishna, Minho Hwang, Vainavi Viswanath, Michael Laskey, Joseph
  Gonzalez, and Ken Goldberg.
\newblock Untangling dense knots by learning task-relevant keypoints.
\newblock In \emph{Conference on Robot Learning}, 2020.

\bibitem[Ha and Song(2021)]{Ha2021FlingBotTU}
Huy Ha and Shuran Song.
\newblock Flingbot: The unreasonable effectiveness of dynamic manipulation for
  cloth unfolding.
\newblock \emph{ArXiv}, abs/2105.03655, 2021.

\bibitem[Handa et~al.(2019)Handa, Wyk, Yang, Liang, Chao, Wan, Birchfield,
  Ratliff, and Fox]{Handa2019DexPilotVT}
Ankur Handa, Karl~Van Wyk, Wei Yang, Jacky Liang, Yu-Wei Chao, Qian Wan, Stan
  Birchfield, Nathan~D. Ratliff, and Dieter Fox.
\newblock Dexpilot: Vision-based teleoperation of dexterous robotic hand-arm
  system.
\newblock \emph{2020 IEEE International Conference on Robotics and Automation
  (ICRA)}, pages 9164--9170, 2019.

\bibitem[He et~al.(2015)He, Zhang, Ren, and Sun]{He2015DeepRL}
Kaiming He, X.~Zhang, Shaoqing Ren, and Jian Sun.
\newblock Deep residual learning for image recognition.
\newblock \emph{2016 IEEE Conference on Computer Vision and Pattern Recognition
  (CVPR)}, pages 770--778, 2015.

\bibitem[Higgins et~al.(2016)Higgins, Matthey, Pal, Burgess, Glorot, Botvinick,
  Mohamed, and Lerchner]{Higgins2016betaVAELB}
Irina Higgins, Lo{\"i}c Matthey, Arka Pal, Christopher~P. Burgess, Xavier
  Glorot, Matthew~M. Botvinick, Shakir Mohamed, and Alexander Lerchner.
\newblock beta-vae: Learning basic visual concepts with a constrained
  variational framework.
\newblock In \emph{International Conference on Learning Representations}, 2016.

\bibitem[Hoque et~al.(2021)Hoque, Balakrishna, Novoseller, Wilcox, Brown, and
  Goldberg]{Hoque2021ThriftyDAggerBN}
Ryan Hoque, Ashwin Balakrishna, Ellen~R. Novoseller, Albert Wilcox, Daniel~S.
  Brown, and Ken Goldberg.
\newblock Thriftydagger: Budget-aware novelty and risk gating for interactive
  imitation learning.
\newblock In \emph{Conference on Robot Learning}, 2021.

\bibitem[James et~al.(2018)James, Bloesch, and
  Davison]{James2018TaskEmbeddedCN}
Stephen James, Michael Bloesch, and Andrew~J. Davison.
\newblock Task-embedded control networks for few-shot imitation learning.
\newblock \emph{ArXiv}, abs/1810.03237, 2018.

\bibitem[Jang et~al.(2022)Jang, Irpan, Khansari, Kappler, Ebert, Lynch, Levine,
  and Finn]{Jang2022BCZZT}
Eric Jang, Alex Irpan, Mohi Khansari, Daniel Kappler, Frederik Ebert, Corey
  Lynch, Sergey Levine, and Chelsea Finn.
\newblock Bc-z: Zero-shot task generalization with robotic imitation learning.
\newblock In \emph{Conference on Robot Learning}, 2022.

\bibitem[Jenness and Wicker(1975)]{jenness_wicker_1975}
R~G Jenness and C~D Wicker.
\newblock Master--slave manipulators and remote maintenance at the oak ridge
  national laboratory, Jan 1975.
\newblock URL \url{https://www.osti.gov/biblio/4179544}.

\bibitem[Johns(2021)]{Johns2021CoarsetoFineIL}
Edward Johns.
\newblock Coarse-to-fine imitation learning: Robot manipulation from a single
  demonstration.
\newblock \emph{2021 IEEE International Conference on Robotics and Automation
  (ICRA)}, pages 4613--4619, 2021.

\bibitem[Ke et~al.(2021)Ke, Wang, Bhattacharjee, Boots, and
  Srinivasa]{ke2021grasping}
Liyiming Ke, Jingqiang Wang, Tapomayukh Bhattacharjee, Byron Boots, and
  Siddhartha Srinivasa.
\newblock Grasping with chopsticks: Combating covariate shift in model-free
  imitation learning for fine manipulation.
\newblock In \emph{International Conference on Robotics and Automation (ICRA)},
  2021.

\bibitem[Kelly et~al.(2018)Kelly, Sidrane, Driggs-Campbell, and
  Kochenderfer]{Kelly2018HGDAggerII}
Michael Kelly, Chelsea Sidrane, K.~Driggs-Campbell, and Mykel~J. Kochenderfer.
\newblock Hg-dagger: Interactive imitation learning with human experts.
\newblock \emph{2019 International Conference on Robotics and Automation
  (ICRA)}, pages 8077--8083, 2018.

\bibitem[Kim et~al.(2021)Kim, Ohmura, and Kuniyoshi]{Kim2021GazeBasedDR}
Heecheol Kim, Yoshiyuki Ohmura, and Yasuo Kuniyoshi.
\newblock Gaze-based dual resolution deep imitation learning for high-precision
  dexterous robot manipulation.
\newblock \emph{IEEE Robotics and Automation Letters}, 6:\penalty0 1630--1637,
  2021.

\bibitem[Kim et~al.(2022)Kim, Ohmura, and Kuniyoshi]{Kim2022RobotPB}
Heecheol Kim, Yoshiyuki Ohmura, and Yasuo Kuniyoshi.
\newblock Robot peels banana with goal-conditioned dual-action deep imitation
  learning.
\newblock \emph{ArXiv}, abs/2203.09749, 2022.

\bibitem[Kingma and Welling(2013)]{Kingma2013AutoEncodingVB}
Diederik~P. Kingma and Max Welling.
\newblock Auto-encoding variational bayes.
\newblock \emph{CoRR}, abs/1312.6114, 2013.

\bibitem[Kroemer et~al.(2015)Kroemer, Daniel, Neumann, van Hoof, and
  Peters]{Kroemer2015TowardsLH}
Oliver Kroemer, Christian Daniel, Gerhard Neumann, Herke van Hoof, and Jan
  Peters.
\newblock Towards learning hierarchical skills for multi-phase manipulation
  tasks.
\newblock \emph{2015 IEEE International Conference on Robotics and Automation
  (ICRA)}, pages 1503--1510, 2015.

\bibitem[Lai et~al.(2022)Lai, Huang, and Gershman]{lai_huang_gershman_2022}
Lucy Lai, Ann~Z Huang, and Samuel~J Gershman.
\newblock Action chunking as policy compression, Sep 2022.
\newblock URL \url{psyarxiv.com/z8yrv}.

\bibitem[Laskey et~al.(2017)Laskey, Lee, Fox, Dragan, and
  Goldberg]{Laskey2017DARTNI}
Michael Laskey, Jonathan Lee, Roy Fox, Anca~D. Dragan, and Ken Goldberg.
\newblock Dart: Noise injection for robust imitation learning.
\newblock In \emph{Conference on Robot Learning}, 2017.

\bibitem[Lee et~al.(2015)Lee, Lu, Gupta, Levine, and Abbeel]{Lee2015LearningFM}
Alex~X. Lee, Henry Lu, Abhishek Gupta, Sergey Levine, and P.~Abbeel.
\newblock Learning force-based manipulation of deformable objects from multiple
  demonstrations.
\newblock \emph{2015 IEEE International Conference on Robotics and Automation
  (ICRA)}, pages 177--184, 2015.

\bibitem[Li(2006)]{Li2006OptimalCF}
Weiwei Li.
\newblock Optimal control for biological movement systems.
\newblock 2006.

\bibitem[Mandlekar et~al.(2021)Mandlekar, Xu, Wong, Nasiriany, Wang, Kulkarni,
  Fei-Fei, Savarese, Zhu, and Mart'in-Mart'in]{Mandlekar2021WhatMI}
Ajay Mandlekar, Danfei Xu, J.~Wong, Soroush Nasiriany, Chen Wang, Rohun
  Kulkarni, Li~Fei-Fei, Silvio Savarese, Yuke Zhu, and Roberto Mart'in-Mart'in.
\newblock What matters in learning from offline human demonstrations for robot
  manipulation.
\newblock In \emph{Conference on Robot Learning}, 2021.

\bibitem[Menda et~al.(2018)Menda, Driggs-Campbell, and
  Kochenderfer]{Menda2018EnsembleDAggerAB}
Kunal Menda, K.~Driggs-Campbell, and Mykel~J. Kochenderfer.
\newblock Ensembledagger: A bayesian approach to safe imitation learning.
\newblock \emph{2019 IEEE/RSJ International Conference on Intelligent Robots
  and Systems (IROS)}, pages 5041--5048, 2018.

\bibitem[Paradis et~al.(2020)Paradis, Hwang, Thananjeyan, Ichnowski, Seita,
  Fer, Low, Gonzalez, and Goldberg]{Paradis2020IntermittentVS}
Samuel Paradis, Minho Hwang, Brijen Thananjeyan, Jeffrey Ichnowski, Daniel
  Seita, Danyal Fer, Thomas Low, Joseph Gonzalez, and Ken Goldberg.
\newblock Intermittent visual servoing: Efficiently learning policies robust to
  instrument changes for high-precision surgical manipulation.
\newblock \emph{2021 IEEE International Conference on Robotics and Automation
  (ICRA)}, pages 7166--7173, 2020.

\bibitem[Pari et~al.(2021)Pari, Muhammad, Arunachalam, and
  Pinto]{pari2021surprising}
Jyothish Pari, Nur Muhammad, Sridhar~Pandian Arunachalam, and Lerrel Pinto.
\newblock The surprising effectiveness of representation learning for visual
  imitation.
\newblock \emph{arXiv preprint arXiv:2112.01511}, 2021.

\bibitem[Pastor et~al.(2009)Pastor, Hoffmann, Asfour, and
  Schaal]{Pastor2009LearningAG}
Peter Pastor, Heiko Hoffmann, Tamim Asfour, and Stefan Schaal.
\newblock Learning and generalization of motor skills by learning from
  demonstration.
\newblock \emph{2009 IEEE International Conference on Robotics and Automation},
  pages 763--768, 2009.

\bibitem[Pomerleau(1988)]{Pomerleau1988ALVINNAA}
Dean~A. Pomerleau.
\newblock Alvinn: An autonomous land vehicle in a neural network.
\newblock In \emph{NIPS}, 1988.

\bibitem[Qin et~al.(2022)Qin, Su, and Wang]{Qin2022FromOH}
Yuzhe Qin, Hao Su, and Xiaolong Wang.
\newblock From one hand to multiple hands: Imitation learning for dexterous
  manipulation from single-camera teleoperation.
\newblock \emph{IEEE Robotics and Automation Letters}, 7:\penalty0
  10873--10881, 2022.

\bibitem[Rahmatizadeh et~al.(2017)Rahmatizadeh, Abolghasemi, B{\"o}l{\"o}ni,
  and Levine]{Rahmatizadeh2017VisionBasedMM}
Rouhollah Rahmatizadeh, Pooya Abolghasemi, Ladislau B{\"o}l{\"o}ni, and Sergey
  Levine.
\newblock Vision-based multi-task manipulation for inexpensive robots using
  end-to-end learning from demonstration.
\newblock \emph{2018 IEEE International Conference on Robotics and Automation
  (ICRA)}, pages 3758--3765, 2017.

\bibitem[Ross et~al.(2010)Ross, Gordon, and Bagnell]{Ross2010ARO}
St{\'e}phane Ross, Geoffrey~J. Gordon, and J.~Andrew Bagnell.
\newblock A reduction of imitation learning and structured prediction to
  no-regret online learning.
\newblock In \emph{International Conference on Artificial Intelligence and
  Statistics}, 2010.

\bibitem[Salehian et~al.(2018)Salehian, Figueroa, and Billard]{Salehian2018AUF}
Seyed Sina~Mirrazavi Salehian, Nadia Figueroa, and Aude Billard.
\newblock A unified framework for coordinated multi-arm motion planning.
\newblock \emph{The International Journal of Robotics Research}, 37:\penalty0
  1205 -- 1232, 2018.

\bibitem[Shafiullah et~al.(2022)Shafiullah, Cui, Altanzaya, and
  Pinto]{Shafiullah2022BehaviorTC}
Nur Muhammad~(Mahi) Shafiullah, Zichen~Jeff Cui, Ariuntuya Altanzaya, and
  Lerrel Pinto.
\newblock Behavior transformers: Cloning k modes with one stone.
\newblock \emph{ArXiv}, abs/2206.11251, 2022.

\bibitem[Shivakumar et~al.(2022)Shivakumar, Viswanath, Gu, Avigal, Kerr,
  Ichnowski, Cheng, Kollar, and Goldberg]{Shivakumar2022SGTM2A}
Kaushik Shivakumar, Vainavi Viswanath, Anrui Gu, Yahav Avigal, Justin Kerr,
  Jeffrey Ichnowski, Richard Cheng, Thomas Kollar, and Ken Goldberg.
\newblock Sgtm 2.0: Autonomously untangling long cables using interactive
  perception.
\newblock \emph{ArXiv}, abs/2209.13706, 2022.

\bibitem[Shridhar et~al.(2021)Shridhar, Manuelli, and
  Fox]{Shridhar2021CLIPortWA}
Mohit Shridhar, Lucas Manuelli, and Dieter Fox.
\newblock Cliport: What and where pathways for robotic manipulation.
\newblock \emph{ArXiv}, abs/2109.12098, 2021.

\bibitem[Shridhar et~al.(2022)Shridhar, Manuelli, and
  Fox]{Shridhar2022PerceiverActorAM}
Mohit Shridhar, Lucas Manuelli, and Dieter Fox.
\newblock Perceiver-actor: A multi-task transformer for robotic manipulation.
\newblock \emph{ArXiv}, abs/2209.05451, 2022.

\bibitem[Sivakumar et~al.(2022)Sivakumar, Shaw, and Pathak]{telekinesis}
Aravind Sivakumar, Kenneth Shaw, and Deepak Pathak.
\newblock Robotic telekinesis: Learning a robotic hand imitator by watching
  humans on youtube.
\newblock \emph{RSS}, 2022.

\bibitem[Smith et~al.(2012)Smith, Karayiannidis, Nalpantidis, Gratal, Qi,
  Dimarogonas, and Kragic]{Smith2012DualAM}
Christian Smith, Yiannis Karayiannidis, Lazaros Nalpantidis, Xavi Gratal, Peng
  Qi, Dimos~V. Dimarogonas, and Danica Kragic.
\newblock Dual arm manipulation - a survey.
\newblock \emph{Robotics Auton. Syst.}, 60:\penalty0 1340--1353, 2012.

\bibitem[Sohn et~al.(2015)Sohn, Lee, and Yan]{Sohn2015LearningSO}
Kihyuk Sohn, Honglak Lee, and Xinchen Yan.
\newblock Learning structured output representation using deep conditional
  generative models.
\newblock In \emph{NIPS}, 2015.

\bibitem[srcteam(2021)]{srcteam_2021}
srcteam.
\newblock Shadow teleoperation system plays jenga, Mar 2021.
\newblock URL \url{https://www.youtube.com/watch?v=7K9brH27jvM}.

\bibitem[srcteam(2022{\natexlab{a}})]{srcteam_2022_1}
srcteam.
\newblock How researchers are using shadow robot's technology, Jun
  2022{\natexlab{a}}.
\newblock URL \url{https://www.youtube.com/watch?v=p36fYIoTD8M}.

\bibitem[srcteam(2022{\natexlab{b}})]{srcteam_2022_2}
srcteam.
\newblock Shadow teleoperation system, Jun 2022{\natexlab{b}}.
\newblock URL \url{https://www.youtube.com/watch?v=cx8eznfDUJA}.

\bibitem[Stepputtis et~al.(2022)Stepputtis, Bandari, Schaal, and
  Amor]{Stepputtis2022ASF}
Simon Stepputtis, Maryam Bandari, Stefan Schaal, and Heni~Ben Amor.
\newblock A system for imitation learning of contact-rich bimanual manipulation
  policies.
\newblock \emph{2022 IEEE/RSJ International Conference on Intelligent Robots
  and Systems (IROS)}, pages 11810--11817, 2022.

\bibitem[Sundaresan et~al.(2021)Sundaresan, Grannen, Thananjeyan, Balakrishna,
  Ichnowski, Novoseller, Hwang, Laskey, Gonzalez, and
  Goldberg]{Sundaresan2021UntanglingDN}
Priya Sundaresan, Jennifer Grannen, Brijen Thananjeyan, Ashwin Balakrishna,
  Jeffrey Ichnowski, Ellen~R. Novoseller, Minho Hwang, Michael Laskey, Joseph
  Gonzalez, and Ken Goldberg.
\newblock Untangling dense non-planar knots by learning manipulation features
  and recovery policies.
\newblock \emph{ArXiv}, abs/2107.08942, 2021.

\bibitem[Swamy et~al.(2022)Swamy, Choudhury, Bagnell, and
  Wu]{Swamy2022CausalIL}
Gokul Swamy, Sanjiban Choudhury, J.~Andrew Bagnell, and Zhiwei~Steven Wu.
\newblock Causal imitation learning under temporally correlated noise.
\newblock In \emph{International Conference on Machine Learning}, 2022.

\bibitem[Tishby and Zaslavsky(2015)]{Tishby2015DeepLA}
Naftali Tishby and Noga Zaslavsky.
\newblock Deep learning and the information bottleneck principle.
\newblock \emph{2015 IEEE Information Theory Workshop (ITW)}, pages 1--5, 2015.

\bibitem[Todorov et~al.(2012)Todorov, Erez, and Tassa]{Todorov2012MuJoCoAP}
Emanuel Todorov, Tom Erez, and Yuval Tassa.
\newblock Mujoco: A physics engine for model-based control.
\newblock \emph{2012 IEEE/RSJ International Conference on Intelligent Robots
  and Systems}, pages 5026--5033, 2012.

\bibitem[Tu et~al.(2021)Tu, Robey, Zhang, and Matni]{Tu2021OnTS}
Stephen Tu, Alexander Robey, Tingnan Zhang, and N.~Matni.
\newblock On the sample complexity of stability constrained imitation learning.
\newblock In \emph{Conference on Learning for Dynamics \& Control}, 2021.

\bibitem[Vaswani et~al.(2017)Vaswani, Shazeer, Parmar, Uszkoreit, Jones, Gomez,
  Kaiser, and Polosukhin]{Vaswani2017AttentionIA}
Ashish Vaswani, Noam~M. Shazeer, Niki Parmar, Jakob Uszkoreit, Llion Jones,
  Aidan~N. Gomez, Lukasz Kaiser, and Illia Polosukhin.
\newblock Attention is all you need.
\newblock \emph{ArXiv}, abs/1706.03762, 2017.

\bibitem[Wiznitzer et~al.()Wiznitzer, Schmitt, and
  Trossen]{Wiznitzer_interbotix_ros_manipulators}
Solomon Wiznitzer, Luke Schmitt, and Matt Trossen.
\newblock interbotix\_ros\_manipulators.
\newblock URL \url{https://github.com/Interbotix/interbotix_ros_manipulators}.

\bibitem[Xie et~al.(2020)Xie, Chowdhury, Kaluza, Zhao, Wong, and
  Yu]{Xie2020DeepIL}
Fan Xie, A.~M. Masum~Bulbul Chowdhury, M.~Clara De~Paolis Kaluza, Linfeng Zhao,
  Lawson L.~S. Wong, and Rose Yu.
\newblock Deep imitation learning for bimanual robotic manipulation.
\newblock \emph{ArXiv}, abs/2010.05134, 2020.

\bibitem[Zeng et~al.(2020)Zeng, Florence, Tompson, Welker, Chien, Attarian,
  Armstrong, Krasin, Duong, Sindhwani, and Lee]{Zeng2020TransporterNR}
Andy Zeng, Peter~R. Florence, Jonathan Tompson, Stefan Welker, Jonathan Chien,
  Maria Attarian, Travis Armstrong, Ivan Krasin, Dan Duong, Vikas Sindhwani,
  and Johnny Lee.
\newblock Transporter networks: Rearranging the visual world for robotic
  manipulation.
\newblock In \emph{Conference on Robot Learning}, 2020.

\bibitem[Zhang et~al.(2017)Zhang, McCarthy, Jow, Lee, Goldberg, and
  Abbeel]{Zhang2017DeepIL}
Tianhao Zhang, Zoe McCarthy, Owen Jow, Dennis Lee, Ken Goldberg, and P.~Abbeel.
\newblock Deep imitation learning for complex manipulation tasks from virtual
  reality teleoperation.
\newblock \emph{2018 IEEE International Conference on Robotics and Automation
  (ICRA)}, pages 1--8, 2017.

\bibitem[Zhou et~al.(2023)Zhou, Kim, Wang, Florence, and Finn]{Zhou2023NeRFIT}
Allan Zhou, Moo~Jin Kim, Lirui Wang, Peter~R. Florence, and Chelsea Finn.
\newblock Nerf in the palm of your hand: Corrective augmentation for robotics
  via novel-view synthesis.
\newblock \emph{ArXiv}, abs/2301.08556, 2023.

\bibitem[Áron Horváth et~al.(2022)Áron Horváth, Ferentzi, Schwartz, Jacobs,
  Meyns, and Köteles]{HORVATH2022}
Áron Horváth, Eszter Ferentzi, Kristóf Schwartz, Nina Jacobs, Pieter Meyns,
  and Ferenc Köteles.
\newblock The measurement of proprioceptive accuracy: A systematic literature
  review.
\newblock \emph{Journal of Sport and Health Science}, 2022.
\newblock ISSN 2095-2546.
\newblock \doi{https://doi.org/10.1016/j.jshs.2022.04.001}.
\newblock URL
  \url{https://www.sciencedirect.com/science/article/pii/S2095254622000473}.

\end{thebibliography}

\clearpage
\newpage

\appendix

\subsection{Comparing \textit{ALOHA} with Prior Teleoperation Setups}
\label{appendix_compare}

\begin{figure*}[t!]
    \centering
    \includegraphics[width=18.7cm,trim={1.3cm 17cm 15cm 3cm},clip]{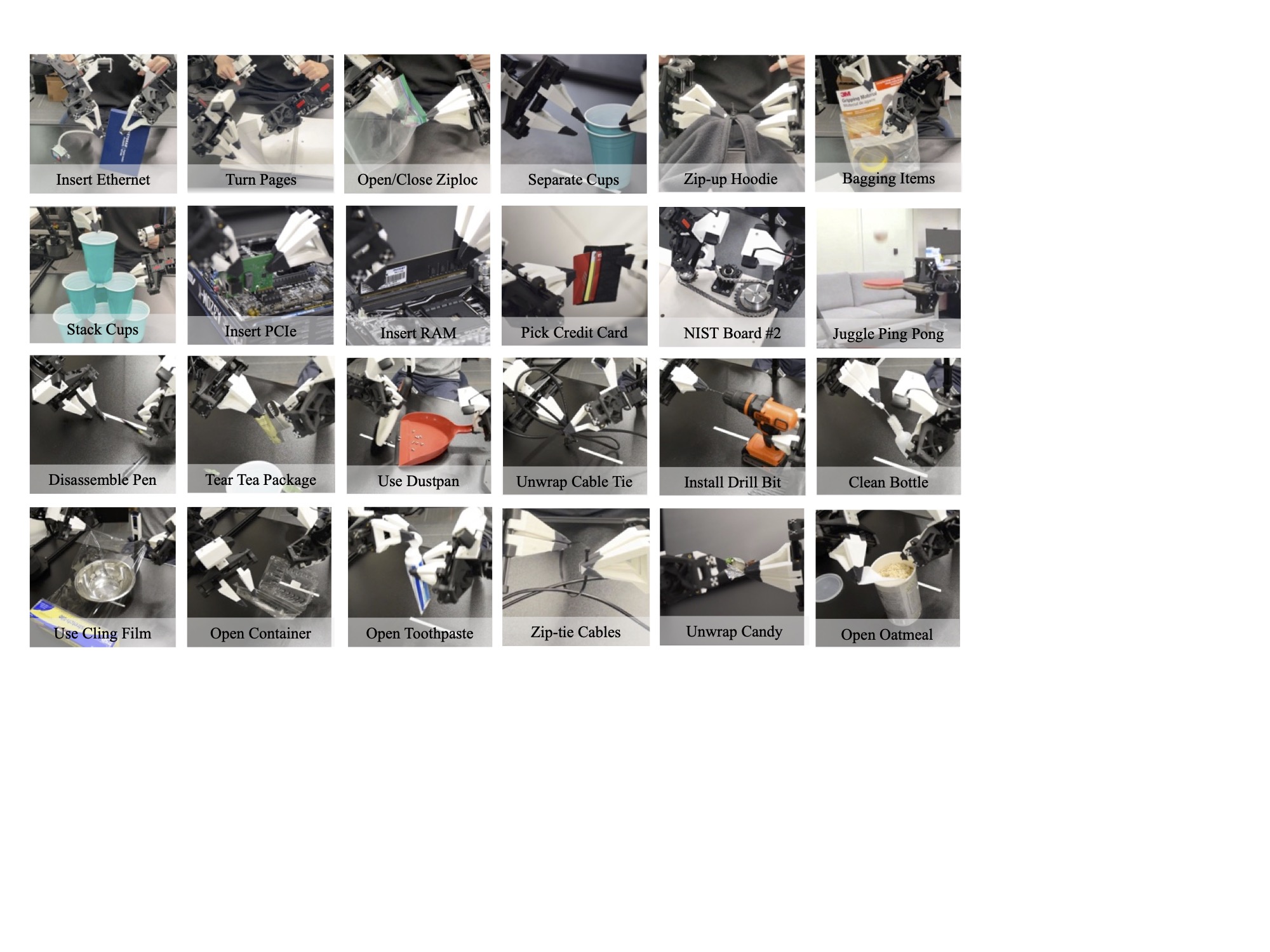}
    \vspace*{-2mm}
    \caption{\small Teleoperation task examples with \textit{ALOHA}. We include videos on the \href{https://tonyzhaozh.github.io/aloha/}{project website}.}
    \label{fig:more_teleop}
    \vspace*{-5mm}
\end{figure*}

In Figure~\ref{fig:more_teleop}, we include more teleoperated tasks that \textit{ALOHA} is capable of.
We stress that all objects are taken directly from the real world without any modification, to demonstrate \textit{ALOHA}'s generality in real life settings.

\textit{ALOHA} exploits the kinesthetic similarity between leader and follower robots by using joint-space mapping for teleoperation.
A leader-follower 
design choice dates back to at least as far as 1953, when Central Research Laboratories built teleoperation systems for handling hazardous material \cite{jenness_wicker_1975}.
More recently, companies like RE2 \cite{youtube_2014} also built highly dexterous teleoperation systems with joint-space mapping.
\textit{ALOHA} is similar to these previous systems, while benefiting significantly from recent advances of low-cost actuators and robot arms. It allows us to achieve similar levels of dexterity with much lower cost, and also without specialized hardware or expert assembly.

Next, we compare the cost of \textit{ALOHA} to recent teleoperation systems.
\textit{DexPilot} \cite{Handa2019DexPilotVT} controls a dexterous hand using image streams of a human hand.
It has 4 calibrated Intel Realsense to capture the point cloud of a human hand, and retarget the pose to an Allegro hand. The Allegro hand is then mounted to a KUKA LBR iiwa7 R800.
DexPilot allows for impressive tasks such as extracting money from a wallet, opening a penut jar, and insertion tasks in NIST board \#1.
We estimate the system cost to be around \$100k with one arm+hand.
More recent works such as Robotic Telekinesis \cite{telekinesis, arunachalam2022holo, Qin2022FromOH} seek to reduce the cost of DexPilot by using a single RGB camera to detect hand pose, and retarget using learning techniques.
While sensing cost is greatly reduced, the cost for robot hand and arm remains high: a dexterous hand has more degrees of freedom and is naturally pricier. Moving the hand around would also require an industrial arm with at least 2kg payload, increasing the price further.
We estimate the cost of these systems to be around \$18k with one arm+hand.
Lastly, the \textit{Shadow Teleoperation System} is a bimanual system for teleoperating two dexterous hands. Both hands are mounted to a UR10 robot, and the hand pose is obtained by either a tracking glove or a haptic glove. This system is the most capable among all aforementioned works, benefitted from its bimanual design. However, it also costs the most, at at least \$400k.
\textit{ALOHA}, on the other hand, is a bimanual setup that costs \$18k (\$20k after adding optional add-ons such as cameras).
Reducing dexterous hands to parallel jaw grippers allows us to use light-weight and low-cost robots, which can be more nimble and require less service.

Finally, we compare the capabilities of \textit{ALOHA} with previous systems.
We choose the most capable system as reference: the \textit{Shadow Teleoperation System} \cite{shadow_robot_2022}, which costs more than 10x of \textit{ALOHA}.
Specifically, we found three demonstration videos \cite{srcteam_2021, srcteam_2022_1, srcteam_2022_2} that contain 15 example use cases of the \textit{Shadow Teleoperation System}, and seek to recreate them using \textit{ALOHA}.
The tasks include playing ``beer pong'', ``jenga,'' and a rubik's cube, using a dustpan and brush, twisting open a water bottle, pouring liquid out, untying velcro cable tie, picking up an egg and a light bulb, inserting and unplugging {USB, RJ45}, using a pipette, writing, twisting open an aluminum case, and in-hand rotation of Baoding balls.
We are able to recreate 14 out of the 15 tasks with similar objects and comparable amount of time. We cannot recreate the Baoding ball in-hand rotation task, as our setup does not have a hand.

\subsection{Example Image Observations}
\label{appendix_example_image}
\begin{figure*}[t!]
    \centering
    \includegraphics[width=0.8\linewidth,trim={0 7.5cm 0 0},clip]{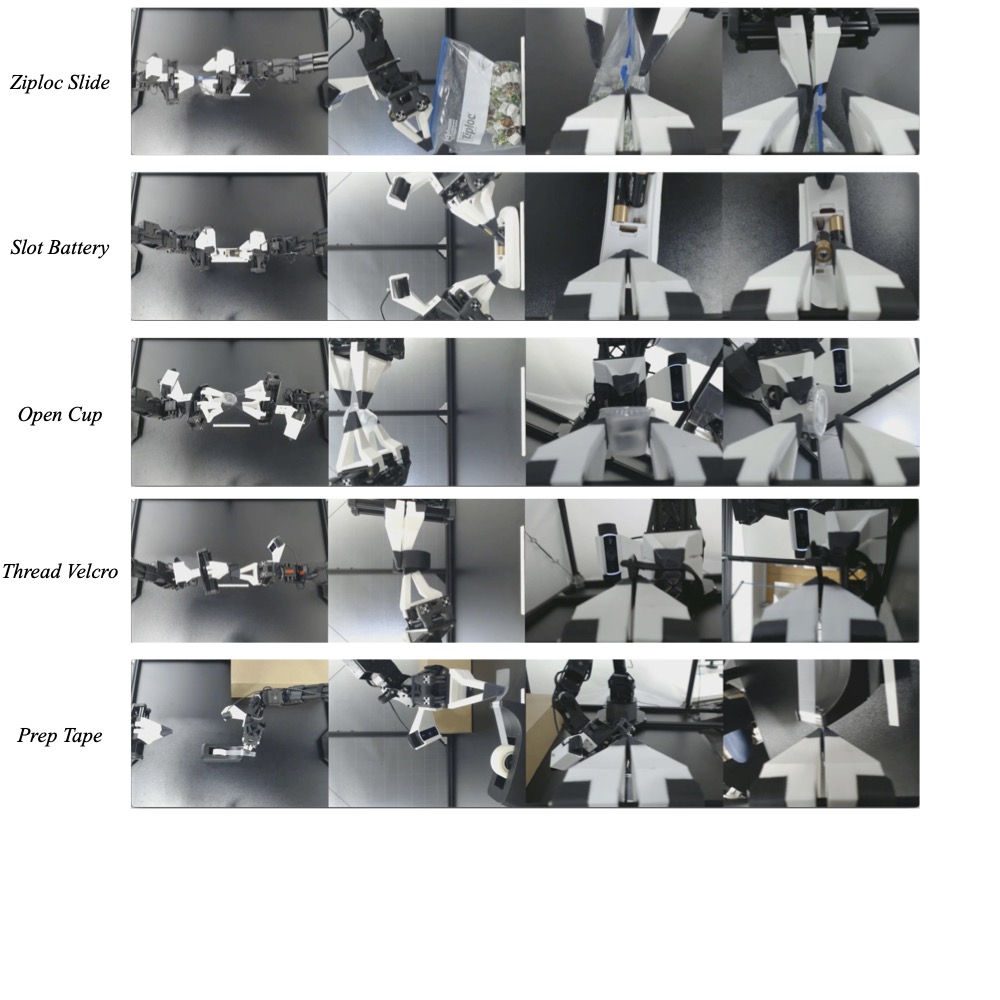}
    \caption{\small Image observation examples for 5 real-world tasks. The 4 columns are [top camera, front camera, left wrist camera, right wrist camera] respectively. We rotate the front camera by 90 degree to capture more vertical space.}
    \label{fig:obs}
\end{figure*}

We include example image observations taken during policy execution time in Figure~\ref{fig:obs}, for each of the 6 real tasks.
From left to right, the 4 images are from top camera, front camera, left wrist, and right wrist respectively.
The top and front cameras are static, while the wrist cameras move with the robots and give detailed views of the gripper.
We also rotate the front camera by 90 degrees to capture more vertical space.
For all cameras, the focal length is fixed with auto-exposure on to adjust for changing lighting conditions. All cameras steam at $480 \times 640$  and 30fps.

\subsection{Detailed Architecture Diagram}
\label{appendix_architecture}
\begin{figure*}[t!]
    \centering
    \includegraphics[width=1\linewidth,trim={0 2cm 0 0},clip]{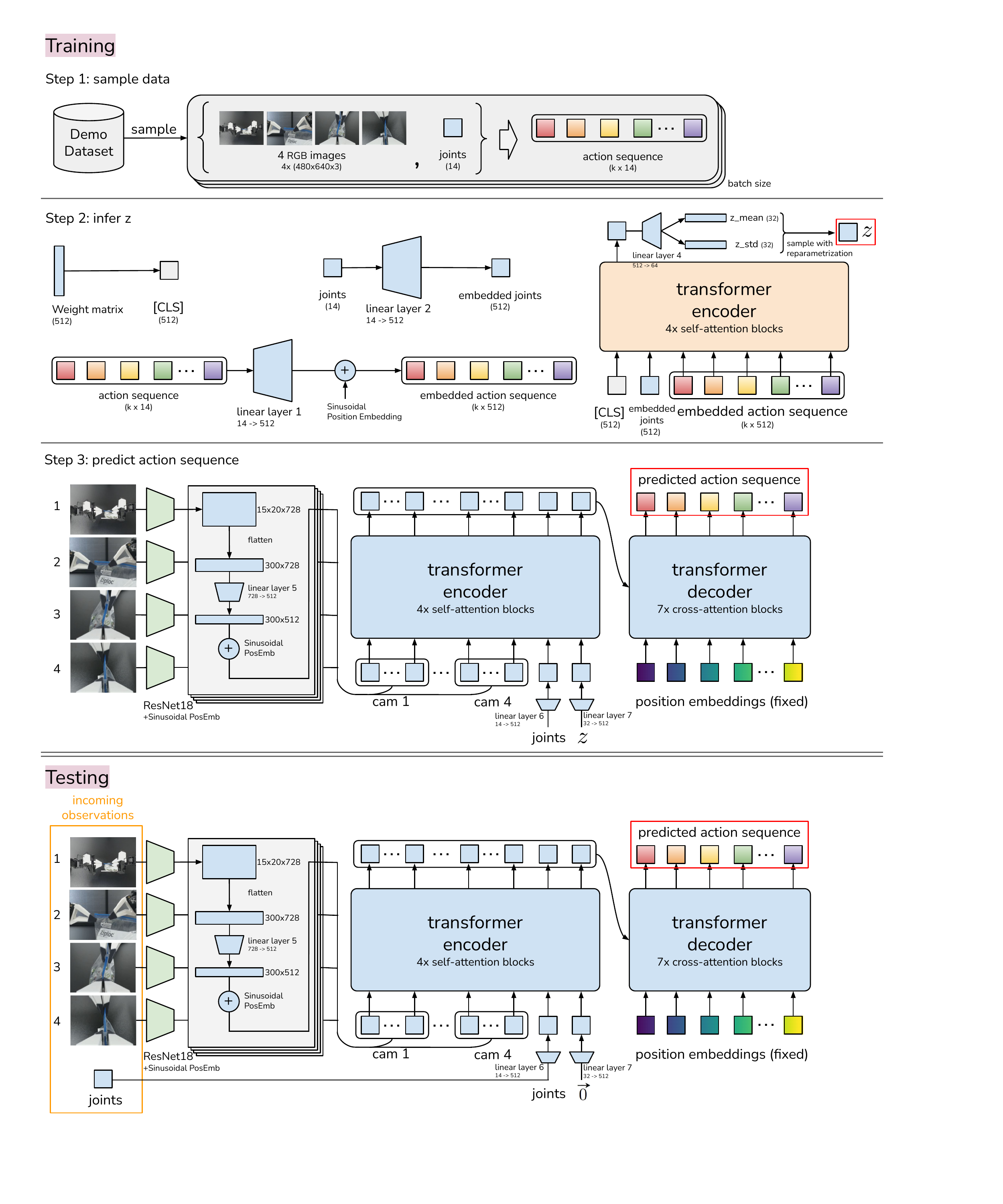}
    \vspace*{-5mm}
    \caption{\small Detail architecture of Action Chunking with Transformers (ACT).}
    \label{fig:detail_archi}
    \vspace*{15mm}
\end{figure*}

We include a more detailed architecture diagram in  Figure~\ref{fig:detail_archi}.
At training time, we first sample tuples of RGB images and joint positions, together with the corresponding action sequence as prediction target (Step 1: sample data).
We then infer style variable $z$ using CVAE encoder shown in yellow (Step 2: infer $z$).
The input to the encoder are 1) the [CLS] token, which consists of learned weights that are randomly initialized, 2) embedded joint positions, which are joint positions projected to the embedding dimension using a linear layer, 3) embedded action sequence, which is the action sequence projected to the embedding dimension using another linear layer.
These inputs form a sequence of $(k+2) \times embedding\_dimension$, and is processed with the transformer encoder.
We only take the first output, which corresponds to the [CLS] token, and use another linear network to predict the mean and variance of $z$'s distribution, parameterizing it as a diagonal Gaussian.
A sample of $z$ is obtained using reparameterization, a standard way to allow back-propagating through the sampling process so the encoder and decoder can be jointly optimized \cite{Kingma2013AutoEncodingVB}. 

Next, we try to obtain the predicted action from CVAE decoder i.e. the policy (Step 3: predict action sequence).
For each of the image observations, it is first processed by a ResNet18 to obtain a feature map, and then flattened to get a sequence of features.
These features are projected to the embedding dimension with a linear layer, and we add a 2D sinusoidal position embedding to perserve the spatial information.
The feature sequence from each camera is then concatenated to be used as input to the transformer encoder. Two additional inputs are joint positions and $z$, which are also projected to the embedding dimension with two linear layers respectively.
The output of the transformer encoder are then used as both ``keys'' and ``values'' in cross attention layers of the transformer decoder, which predicts action sequence given encoder output.
The ``queries'' are fixed sinusoidal embeddings for the first layer.

At test time, the CVAE encoder (shown in yellow) is discarded and the CVAE decoder is used as the policy.
The incoming observations (images and joints) are fed into the model in the same way as during training. The only difference is in $z$, which represents the ``style'' of the action sequence we want to elicit from the policy. 
We simply set $z$ to a zero vector, which is the mean of the unit Gaussian prior used during training.
Thus given an observation, the output of the policy is always deterministic, benefiting policy evaluation.

\begin{figure*}[tbh]
    \centering
    \includegraphics[width=0.2\linewidth,trim={0 4.5cm 0 0},clip]{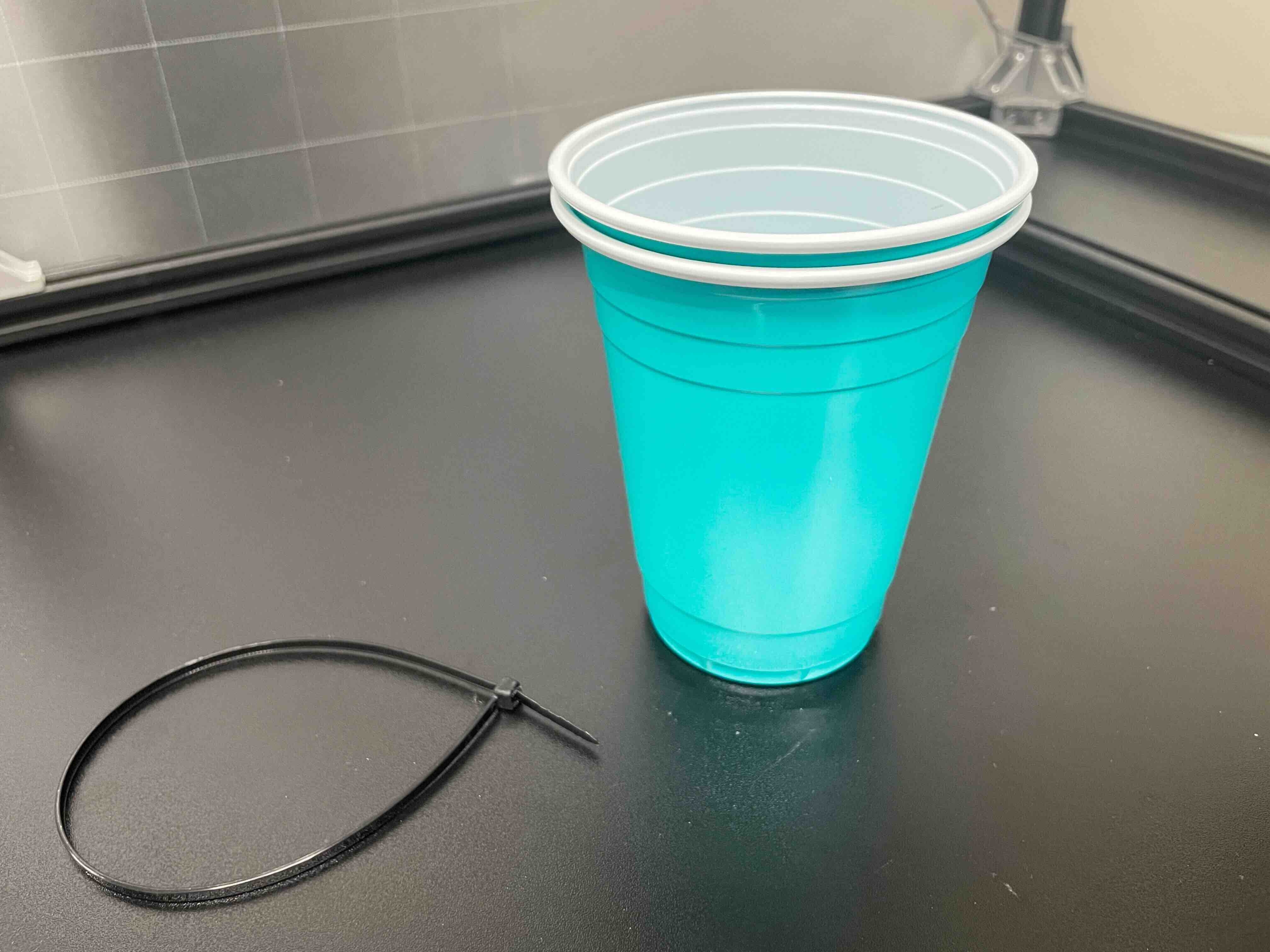}
    \caption{\small The cable tie and cups for user study.}
    \label{fig:user_study_obj}
\end{figure*}

\subsection{Experiment Details and Hyperparameters}
\label{appendix_hparams}
We carefully tune the baselines and include the hyperparameters used in Table~\ref{table:hparam_act}, \ref{table:hparam_byol}, \ref{table:hparam_bet}, \ref{table:hparam_vinn}, \ref{table:hparam_rt1}.
For BeT, we found that increasing history length from 10 (as in original paper) to 100 greatly improves the performance. Large hidden dimension also generally helps.
For VINN, the k used when retrieving nearest neighbor is adaptively chosen with the lowest validation loss, same as the original paper. We also found that using joint position differences in addition to visual feature similarity improves performance when there is no action chunking, in which case we have state weight = 10 when retrieving actions. However, we found this to hurt performance with action chunking and thus set state weight to 0 for action chunking experiments.

\subsection{User Study Details}
\label{appendix_user_study}

We conduct the user study with 6 participants, recruited from computer
science graduate students, with 4 men and 2 women aged 22-25.
3 of the participants had experience teleoperating robots with a VR controller, and the other 3 has no prior experience teleoperating.
None of the participants used \textit{ALOHA} before.
To implement the 5Hz version of \textit{ALOHA}, we read from the leader robot at 5Hz, interpolate in the joint space, and send the interpolated positions to the robot at 50Hz.
We choose tasks that emphasizes high-precision and close-loop visual feedback. We include images of the objects used in Figure~\ref{fig:user_study_obj}.
For threading zip cable tie, the hole measures 4mm x 1.5mm, and the cable tie measures 0.8mm x 3.5mm with a pointy tip. It is initially lying flat on the table, and the operator needs to pick it up with one gripper, grasp the other end mid-air, then coordinate both hands to insert one end of the cable tie into the hole on the other end.
For unstacking cup, we use two single-use plastic cups that has 2.5mm clearance between them when stacked. The teleoperator need to grasp the edge of upper cup, then either shake to separate or use the help from the other gripper.
During the user study, we randomize the order in which operators attempt each task, and whether they use 50Hz or 5Hz controller first.
We also randomize the initial position of the object randomly around the table center.
For each setting, the operator has 2 minutes to adapt, followed by 3 consecutive attempts of the task with duration recorded.

\subsection{Limitations}
\label{appendix_limitations}
We now discuss limitations of the ALOHA hardware and the policy learning with ACT.

\noindent \textbf{Hardware Limitations.} On the hardware front, \textit{ALOHA} struggles with tasks that require multiple fingers from both hands, for example opening child-proof pill bottles with a push tab. To open the bottle, one hand needs to hold the bottle and pushes down on the push tab, with the other hand twisting the lid open.
\textit{ALOHA} also struggles with tasks that require high amount of forces, for example lifting heavy objects, twisting open a sealed bottle of water, or opening markers caps that are tightly pressed together.
This is because the low-cost motors cannot generate enough torque to support these manipulations. 
Tasks that requires finger nails are also difficult for \textit{ALOHA}, even though we design the grippers to be thin on the edge. For example, we are not able to lift the edge of packing tape when it is taped onto itself,
or opening aluminum soda cans.

\noindent \textbf{Policy Learning Limitations.} On the software front, we report all 2 tasks that we attempted where ACT failed to learn the behavior.
The first one is unwrapping candies. The steps involves picking up the candy from the table, pull on both ends of it, and pry open the wrapper to expose the candy.
We collected 50 demonstrations to train the ACT policy. In our preliminary evaluation with 10 trials, the policy picks up the candy 10/10, pulls on both ends 8/10, while unwraps the candy 0/10.
We attribute the failure to the difficulty of perception and lack of data. Specifically, after pulling the candy on both sides, the seam for prying open the candy wrapper could appear anywhere around the candy. During demonstration collection, it is difficult even for human to discern. The operator needs to judge by looking at the graphics printed on the wrapper and find the discontinuity.
We constantly observe the policy trying to peel at places where the seam does not exist.
To better track the progress, we attempted another evaluation where we give 10 trials for each candy, and repeat this for 5 candies. For this protocol, our policy successfully unwraps 3/5 candies.

Another task that ACT struggles with is opening a small ziploc bag laying flat on the table. The right gripper needs to first pick it up, adjust it so that the left gripper can grasp firmly on the pulling region, followed by the right hand grasping the other side of the pulling region, and pull it open.
Our policy trained with 50 demonstrations can consistently pick up the bag, while having difficulties performing the following 3 mid-air manipulation steps.
We hypothesize that the bag is hard to perceive, and in addition, small differences in the pick up position can affect how the bag deforms, and result in large differences in where the pulling region ends up.
We believe that pretraining, more data, and better perception are promising directions to tackle these extremely difficult tasks.

\begin{table*}[tbh!]
\centering
\setlength{\tabcolsep}{32pt}
\begin{tabular}{ll}
\toprule
learning rate & 1e-5 \\
batch size & 8 \\
\# encoder layers & 4 \\
\# decoder layers & 7 \\
feedforward dimension & 3200 \\
hidden dimension & 512 \\
\# heads & 8 \\
chunk size & 100 \\
beta & 10 \\
dropout & 0.1 \\

\bottomrule
\end{tabular}
\caption{\small Hyperparameters of ACT.}
\label{table:hparam_act}
\end{table*}

\begin{table*}[tbh!]
\centering
\setlength{\tabcolsep}{32pt}
\begin{tabular}{ll}
\toprule
learning rate & 3e-4 \\
batch size & 128\\
epochs & 100 \\
momentum & 0.9 \\
weight decay & 1.5e-6 \\

\bottomrule
\end{tabular}
\caption{\small Hyperparameters of BYOL, the feature extractor for VINN and BeT.}
\label{table:hparam_byol}
\end{table*}

\begin{table*}[tbh!]
\centering
\setlength{\tabcolsep}{32pt}
\begin{tabular}{ll}
\toprule
learning rate & 1e-4 \\
batch size & 64 \\
\# layers & 6 \\
\# heads & 6 \\
hidden dimension & 768 \\
history length & 100 \\
weight decay & 0.1 \\
offset loss scale & 1000 \\
focal loss gamma & 2 \\
dropout & 0.1 \\
discretizer \#bins & 64 \\

\bottomrule
\end{tabular}
\caption{\small Hyperparameters of BeT.}
\label{table:hparam_bet}
\end{table*}

\begin{table*}[tbh!]
\centering
\setlength{\tabcolsep}{32pt}
\begin{tabular}{ll}
\toprule
k (nearest neighbour) & adaptive\\
state weight & 0 or 10 \\

\bottomrule
\end{tabular}
\caption{\small Hyperparameters of VINN.}
\label{table:hparam_vinn}
\end{table*}

\begin{table*}[tbh!]
\centering
\setlength{\tabcolsep}{32pt}
\begin{tabular}{ll}
\toprule
learning rate & 1e-5 \\
batch size & 2 \\
ViT dim head & 32 \\
ViT window size & 7 \\
ViT mbconv expansion rate & 4 \\
ViT mbconv shrinkage rate & 0.25 \\
ViT dropout & 0.1 \\
RT-1 depth & 6 \\
RT-1 heads & 8 \\
RT-1 dim head & 64 \\
RT-1 action bins & 256 \\
RT-1 cond drop prob & 0.2 \\
RT-1 token learner num output tokens & 8 \\
weight decay & 0 \\
history length & 6 \\

\bottomrule
\end{tabular}
\caption{\small Hyperparameters of RT-1.}
\label{table:hparam_rt1}
\end{table*}

\end{document}